\documentclass{article} 
\usepackage{iclr2025_conference,times}

\usepackage{amsmath,amsfonts,bm}

\def\eqref#1{equation~\ref{#1}}

\def\1{\bm{1}}

\DeclareMathAlphabet{\mathsfit}{\encodingdefault}{\sfdefault}{m}{sl}
\SetMathAlphabet{\mathsfit}{bold}{\encodingdefault}{\sfdefault}{bx}{n}

\usepackage{hyperref}
\usepackage{url}

\usepackage{booktabs}       
\usepackage{amsfonts}       
\usepackage{nicefrac}       
\usepackage{microtype}      
\usepackage{xcolor}         
\usepackage{graphicx}
\usepackage{amsmath}
\usepackage{amssymb}
\usepackage{mathtools}
\usepackage{amsthm}
\usepackage{placeins}

\usepackage{bm}
\usepackage{algorithm}
\usepackage{algpseudocode}
\usepackage{subfigure}
\usepackage{wrapfig}
\usepackage[capitalize,noabbrev]{cleveref}

\usepackage{multirow}
\usepackage{xspace}
\usepackage{caption}
\usepackage{thm-restate}
\usepackage{footmisc}

\newcommand{\ie}{\textit{i}.\textit{e}.\xspace}
\newcommand{\eg}{\textit{e}.\textit{g}.\xspace}
\newcommand{\etc}{\textit{etc\@\xspace}}
\newcommand{\B}{\bfseries}

\newcommand{\x}{\bm{x}}
\newcommand{\y}{\bm{y}}
\newcommand{\z}{\mathbf{z}}

\newcommand{\I}{\bm{I}}

\newcommand{\Vc}{\bm{c}}

\newcommand\myldots{\ifmmode\ldots\else\makebox[0.5em][c]{.\hfil.\hfil.}\thinspace\fi}
\newcommand\mycdots{\ifmmode\cdots\else\makebox[0.5em][c]{.\hfil.\hfil.}\thinspace\fi}

\theoremstyle{plain}

\theoremstyle{definition}

\theoremstyle{remark}

\title{Enhanced Diffusion Sampling via \\ Extrapolation with Multiple ODE Solutions}

\author{Jinyoung Choi$^{1}$, Junoh Kang$^{1}$ \& Bohyung Han$^{1,2}$ \\ Computer Vision Lab., 
$^{1}$ECE \& $^{2}$IPAI, Seoul National University, Korea\\
\texttt{\{jin0.choi, junoh.kang, bhhan\}@snu.ac.kr} \\
}

\iclrfinalcopy 
\begin{document}

\maketitle


\begin{abstract}
Diffusion probabilistic models (DPMs), while effective in generating high-quality samples, often suffer from high computational costs due to their iterative sampling process.
To address this, we propose an enhanced ODE-based sampling method for DPMs inspired by Richardson extrapolation, which reduces numerical error and improves convergence rates.
Our method, RX-DPM, leverages multiple ODE solutions at intermediate time steps to extrapolate the denoised prediction in DPMs.
This significantly enhances the accuracy of estimations for the final sample while maintaining the number of function evaluations (NFEs).
Unlike standard Richardson extrapolation, which assumes uniform discretization of the time grid, we develop a more general formulation tailored to arbitrary time step scheduling, guided by local truncation error derived from a baseline sampling method.
The simplicity of our approach facilitates accurate estimation of numerical solutions without significant computational overhead, and allows for seamless and convenient integration into various DPMs and solvers. 
Additionally, RX-DPM provides explicit error estimates, effectively demonstrating the faster convergence as the leading error term's order increases.
Through a series of experiments, we show that the proposed method improves the quality of generated samples without requiring additional sampling iterations.

\end{abstract}
\section{Introduction}
\label{sec:introduction}

Diffusion probabilistic models (DPMs) have emerged as a powerful framework for generating high-quality samples in a wide range of applications and domains for images~\citep{ho2020denoising,song2021scorebased,dhariwal2021diffusion,rombach2022high}, videos~\citep{ho2022video,singer2022make,zhou2022magic,wang2023modelscope}, 3D shapes~\citep{zeng2022lion}, \etc. 
While DPMs demonstrate impressive performance in data fidelity and diversity, they also have limitations, particularly their computational inefficiency due to the sequential nature of sampling. 
Addressing this issue is crucial for enhancing the usability of DPMs in real-world scenarios, where time constraints are critical for practical deployment.

The generation process of DPMs can be formulated as a problem of finding solutions to SDEs or ODEs~\citep{song2021scorebased}, where the truncation errors of the numerical solutions are highly correlated to the quality of the generated samples. 
To enhance the quality of these samples, it is essential to reduce truncation errors, which can be achieved by adopting advanced solvers or numerical techniques that improve the accuracy of numerical estimations. 
In this context, we aim to lower truncation errors by applying numerical extrapolation to existing sampling methods for DPMs.
The key ingredient of the proposed method is Richardson extrapolation, a proven and widely used technique in the mathematical modeling of physical problems such as fluid dynamics and heat transfer, which demand high computational resources.
While numerous variants and strategies have been studied~\citep{richards1997completed,botchev2009numerical,zlatev2010stability}, its application to DPMs remains unexplored.
The method uses a simple linear combination of multiple numerical estimates from different resolutions of a grid to approximate the ideal solution, where the estimates are expected to converge as the resolution becomes finer, ultimately reaching the target value in the limit.

\begin{wrapfigure}{r}{0.44\textwidth}
    \vspace{1mm}
    \centering
    \includegraphics[width=0.98\linewidth]{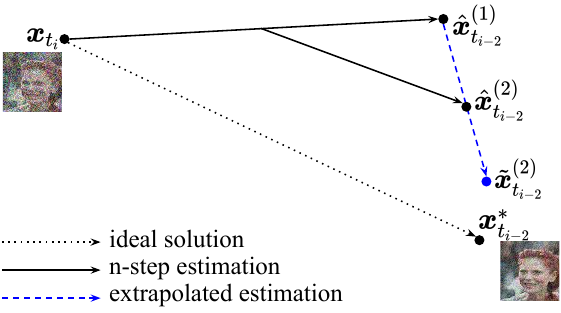} 
    \caption{
        Application of the proposed extrapolation on two denoising steps ($k=2$) with time steps of $[t_{i},t_{i-1},t_{i-2}]$.
        $\hat\x_{t_{i-2}}^{(n)}$ denotes that $n$ steps are used by the baseline sampler within the same interval.
        $\tilde\x_{t_{i-2}}^{(2)}$ represents the extrapolated estimation using two ODE solutions at $t_{i-2}$, $\hat\x_{t_{i-2}}^{(1)}$ and $\hat\x_{t_{i-2}}^{(2)}$.
    }
    \label{fig:extrapolation}
\end{wrapfigure}
We propose an extrapolation algorithm that is applied repeatedly every $k$ denoising steps of an ODE-based sampling method to improve the accuracy of intermediate denoising steps. 
This is achieved by utilizing an additional ODE solution, which is estimated by a single step over an interval of $k$ time steps. 
\cref{fig:extrapolation} illustrates this concept with $k=2$ on time steps $[t_{i},t_{i-1},t_{i-2}]$, which forms a unit block of our extrapolation-based sampling. 
Two ODE solutions---single-step and two-step estimations at $t_{i-2}$ from $t_{i}$---can be leveraged to achieve an approximation closer to the ideal solution $\x_{t_{i-2}}^*$, which is unknown.

The standard Richardson extrapolation assumes a uniform discretization of the time grid.
However, the uniform grid of denoising time steps might be suboptimal for DPMs; a smaller time interval near the clean sample is often more beneficial~\citep{karras2022elucidating,song2021denoising} given the same number of steps.
Considering such characteristics of DPMs and the advantages offered by specific time step discretizations, we propose an algorithm applicable to arbitrary schedules, with coefficients determined by the chosen configuration.  
We observe that our grid-aware approach yields better performance than conventional methods.

Although there exist other methods applying extrapolation techniques to diffusion models, their usages of extrapolation are somewhat different from ours. 
For example, \citet{zhang2023fast,zhang2023lookahead} utilize estimations from earlier steps to improve the estimation at the time step, $t_i$, whereas our approach adopts two denoised estimations at the same time step, $t_i$, to enhance accuracy at $t_i$. 
In addition, the main building block of our approach, Richardson extrapolation, is proven to enhance numerical accuracy and provides an explicit estimate of the error, which allows for a clear understanding of the convergence behavior. 
Furthermore, the implementation of our algorithm is simple and cost-effective because it requires no additional network evaluations and insignificant computational overhead to perform the extrapolation. 
We refer to the proposed sampling algorithm as RX-DPM. 

Our main contributions are summarized below:
\begin{itemize}
    \item We introduce an improved diffusion sampler, RX-DPM, inspired by Richardson extrapolation, which effectively increases the order of accuracy of existing ODE-based samplers without increasing NFEs. 
    \item We systematically develop an algorithm for general DPM solvers with arbitrary time step scheduling starting from the derivation of a truncation error of the Euler method on a non-uniform grid.
    \item Our experiments across various well-known baselines demonstrate that RX-DPM exhibits strong generalization performance and high practicality, regardless of ODE designs, model architectures, and base samplers.
\end{itemize}


\section{Related work}
\label{sec:related_work}
\vspace{-1mm}

There exists a substantial body of research that seeks to reduce the computational burden of DPMs while maintaining their performance. 
One approach in this direction involves exploring alternative modeling strategies for the reverse process.
For example, the networks in \citet{salimans2022progressive, song2023consistency, kim2024ctm} learn alternative objectives through knowledge distillation, using the outputs obtained by iterative inferences of the pretrained (teacher) networks.
On the other hand, \citet{bao2022estimating} model a more accurate reverse distribution by incorporating the optimal covariance, while \citet{xiao2022DDGAN, kang2024ogdm} implicitly estimate the precise reverse distribution by utilizing GAN components.

While the aforementioned methods require model training, there is also a training-free approach that interprets the generation process of a diffusion model as solving an ODE or SDE~\citep{song2021scorebased,karras2022elucidating}.
For instance, DDIM~\citep{song2021denoising} proposes to skip intermediate time steps, which is equivalent to solving an ODE using the Euler method with a large step size.
To further improve sampling quality, a large volume of research~\citep{karras2022elucidating,dockhorn2022genie,liu2022pseudo,zhang2023fast,lu2022fastode,lu2023dpm} simply applies classical higher-order solvers or customizes them for diffusion models.
Specifically, \citet{karras2022elucidating} adopt the second-order Heun's method~\citep{suli2003introduction} and \citet{dockhorn2022genie} apply the second-order Taylor expansion.
In addtion, \citet{liu2022pseudo} propose a pseudo-numerical solver, which combines classical high-order numerical methods and DDIM.

Building on ODE-based sampling methods, various numerical techniques have been employed to further improve the sample quality. 
For example, IIA~\citep{zhang2024iia} optimizes the coefficients of certain quantities to approximate the fine-grained integration. 
LA-DPM~\citep{zhang2023lookahead} and DEIS~\citep{zhang2023fast} adopt extrapolation techniques as our method, but with notable differences. 
Specifically, LA-DPM linearly extrapolates the previous and current predictions of the solution at $t=0$---the original data manifold---while DEIS uses high-order polynomial extrapolation on the noise prediction function.
The key distinction of our approach is that, while they adopt score (noise) predictions obtained at different time steps for each extrapolation, we utilize multiple denoised outputs obtained at the same time step.

\section{Preliminaries}
\label{sec:prelim}

\subsection{Diffusion probabilistic models as solving an ODE}
\label{subsec:diffusion}
For $\bm{p}_0 = \bm{p}_\text{data}$ and $\x \in \mathbb{R}^d$, \citet{karras2022elucidating} defines a marginal distribution at $t$ as 
\begin{align}
    \bm{p}_t(\x) = s(t)^{-d} \bm{p}(\x / s(t);\sigma(t)),
\end{align}
where $\bm{p}(\x;\sigma) = \bm{p}_\text{data} * \mathcal{N}(\bm{0}, \sigma(t)^2 \I)$, and $s(t)$ and $\sigma(t)$ are non-negative functions satisfying $s(0)=1$, $\sigma(0)=0$, and $\lim_{t \rightarrow \infty}\frac{\sigma(t)}{s(t)} = \infty$.
The probability flow ODE,
\begin{align}
    d\x = [\dot{s}(t)/s(t) - s(t)^2 \dot{\sigma}(t)\sigma(t) \nabla_{\x} \log p( \x / s(t); \sigma(t))] dt, 
    \quad \x(T) \sim \bm{p}_T(\x),
    \label{eq:general_ode}
\end{align}
matches the marginal distribution.
By adopting the specific choices, $s(t)=1$ and $\sigma(t)=t$ as in~\citet{karras2022elucidating}, \cref{eq:general_ode} is reduced as follows:
\begin{align}
    d\x = - t \nabla_{\x} \log \bm{p}( \x ; t) dt, \quad \x(T) \sim \bm{p}_T(\x).
    \label{eq:edm_ode}
\end{align}

Diffusion models now learn the score function $\nabla_{\x} \log \bm{p}( \x ; t)$, which is the only unknown component in the equation.
For sufficiently large $T$, the marginal distribution $\bm{p}_T(\x)$ can be approximated by $\mathcal{N}(\x;\bm{0},T^2 \I)$ and the generation process is equivalent to solving for $\x(0)$ using \cref{eq:edm_ode} with the boundary condition, $\x(T) \sim \mathcal{N}(\bm{0},T^2\I)$.
Since the analytic solution of \cref{eq:edm_ode} cannot be expressed in a closed form, numerical methods are used to solve the ODE.
Given the time step scheduling, $0 = t_0 < t_1 < \ldots < t_{N} = T$, the solution is given by
\begin{align}
    \x(0) 
    &= \x(T) + \int_{T}^{0} - t \nabla_{\x} \log \bm{p}( \x(t) ; t) dt \\
    &= \x(T) + \sum_{i=N}^{1} \int_{t_i}^{t_{i-1}} - t \nabla_{\x} \log \bm{p}( \x(t) ; t) dt,
\end{align}
where each integration from $t_{i}$ to $t_{i-1}$ can be approximated by ODE solvers such as the Euler method or Heun's method.

\subsection{Richardson extrapolation}
\label{subsec:richardson_extrapolation}

Let the exact and numerical solutions at $t=0$ be $V^*$ and $V(h)$, respectively, where $h$~($0<h<1$) denotes the step size. 
If $V^* = \lim_{h \rightarrow 0 } V(h)$ and the order of truncation error is known, Richardson extrapolation~\citep{richardson1911extra} identifies a faster converging sequence, $\tilde{V}(h)$.
For instance, $V(h)$ with a truncation error in the order of $O(h^p)$ is expressed by
\begin{align}
    V^* = V(h) + c h^p + O(h^q)
    \label{eq:ric_extra1}
\end{align}
for $0 < p < q$ and $c \neq 0$.
Then, for a fixed constant $k>1$,
\begin{align}
    V^* = V(h / k) + \frac{c}{k^p} h^p + O(h^q).
    \label{eq:ric_extra2}
\end{align}
From \cref{eq:ric_extra1,eq:ric_extra2}, eliminating the $h^p$ terms, we obtain the solution
\begin{align}
    \tilde{V}(h,k) = \cfrac{k^p V(h/k) - V(h)}{k^p -1},
    \label{eq:ric_extra_sol}
\end{align}
which has a truncation error of $O(h^q)$, asymptotically smaller than $O(h^p)$.


\section{RX-DPM}
\label{sec:rx-dpm}
Before discussing the proposed method, we first outline the algorithmic development process for the most simplified problem and then explore an extension to a general DPM solver.
The application of our method, RX-DPM, to a specific solver will be referred to as RX-[\textit{SolverName}].

\subsection{Truncation error of Euler method on non-uniform grid}
\label{subsec:simple_euler}

We now derive the truncation error formula for the Euler method on a non-uniform grid, based on the local truncation error, which results from a single iteration. 
For intuitive clarity, we consider a one-dimensional ODE of the form
\[
dx = f(x, t) dt,
\]
where $f$ is a smooth function. 
Suppose that the numerical solution is obtained using the Euler method with the discretization points $[t_i, t_{i-1}, \dots, t_{i-k}]$ in a reverse time order, given the initial condition $x(t_i) = x_{t_i}$.
From now on, we denote $\hat{x}_{t_{j}}^{(n)}$ as the numerical solution at $t_j$ obtained by $n$ iterations and $x_{t_{j}}^{*}$ as the exact solution at $t_j$.
Given $h=t_i-t_{i-k}$ and $\lambda_j = \frac{1}{h}(t_{i-j+1}-t_{i-j})$ for $j=1, \dots, k$, the local truncation error formula of the one-step Euler method, derived from the Taylor expansion, is expressed as
\begin{align}
    \hat{x}_{t_{i-1}}^{(1)} 
    = x_{t_i} -\lambda_1 h f(x_{t_{i}};t_{i})
    = x_{t_{i-1}}^{*} - \frac{1}{2} x_{t_{i}}'' \lambda_1^2 h^2 + O(h^3).
    \label{eq:Euler_error}
\end{align}
Then, the truncation error of the two-step numerical solution is derived as
\begin{align}
    \hat x^{(2)}_{t_{i-2}} 
    &= \hat x^{(1)}_{t_{i-1}} - \lambda_2 h f(\hat x^{(1)}_{t_{i-1}}) \\
    &= x_{t_{i-1}}^* - \frac{1}{2}x_{t_i}^{''}\lambda_1^2h^2 + O(h^3) - \lambda_2hf(\hat x^{(1)}_{t_{i-1}}) \\ 
    &= x_{t_{i-1}}^* - \lambda_2hf(x_{t_{i-1}}^*) - \frac{1}{2}x_{t_i}^{''}\lambda_1^2h^2 + O(h^3) - \lambda_2hf(\hat x^{(1)}_{t_{i-1}}) + \lambda_2hf(x_{t_{i-1}}^*) \\
    &= x_{t_{i-2}}^* - \frac{1}{2}x_{t_{i-1}}^{*''}\lambda_2^2h^2 - \frac{1}{2}x_{t_i}^{''}\lambda_1^2h^2 + O(h^3) \quad (\because 
    \text{\cref{eq:Euler_error} and } f \text{ is smooth}) \\
    &= x_{t_{i-2}}^* - \frac{1}{2}x_{t_i}^{''}(\lambda_1^2 + \lambda_2^2) h^2 + O(h^3) \quad (\because f \text{ is smooth}).
\end{align}

Inductively, we can obtain the truncation error for the $k$-step solution as 
\begin{align}
    \hat x^{(k)}_{t_{i-k}} 
    = x_{t_{i-k}}^* - \frac{1}{2}x_{t_i}^{''} \sum_{j=1}^{k} \lambda_j^2 h^2 + O(h^3),
    \label{eq:truncation_k}
\end{align}
which approximates $x_{t_{i-k}}^*$ with a truncation error of $O(h^2)$. 

\subsection{RX-Euler}
\label{subsec:rx-euler}
We now describe RX-Euler performing extrapolation every $k$ steps on the Euler method.
Extrapolation is executed as a linear combination of two different numerical solutions $\hat{\x}_{t_{i-k}}^{(1)}$ and $\hat{\x}_{t_{i-k}}^{(k)}$ obtained by the Euler solver over a single step on the grid $[t_i, t_{i-k}]$ and $k$ steps on the grid $[t_i, t_{i-1}, \dots, t_{i-k}]$, respectively. 
To calculate coefficients for extrapolation, we use the truncation error derived in \cref{subsec:simple_euler}, which can be also applied to \cref{eq:edm_ode} in \cref{subsec:diffusion}, as the ideal score function is considered smooth; its derivative is Lipschitz continuous, referring to the equation in Appendix B.3 of \citet{karras2022elucidating}.
From \cref{eq:Euler_error,eq:truncation_k}, we derive the expressions for \(\hat{\x}_{t_{i-k}}^{(1)}\) and \(\hat{\x}_{t_{i-k}}^{(k)}\) for a constant $\Vc$ as follows:
\vspace{-2mm}
\begin{align}
    \hat \x^{(1)}_{t_{i-k}} &= \x_{t_{i-k}}^* + \Vc h^2 + O(h^3) 
    \label{eq:euler1} \\
    \hat \x^{(k)}_{t_{i-k}} 
    &= \x_{t_{i-k}}^* + \Vc\sum_{j=1}^{k} \lambda_j^2 h^2 + O(h^3). \label{eq:euler2}
\end{align}
Then, by solving the linear system of \cref{eq:euler1,eq:euler2}, we approximate $\x_{t_{i-k}}^*$ through the following extrapolation:
\begin{align}
\tilde{\x}_{t_{i-k}}^{(k)} = \frac{\hat \x^{(k)}_{t_{i-k}} - \sum_{j=1}^k \lambda_j^2 \hat \x^{(1)}_{t_{i-k}}}{1-\sum_{j=1}^k \lambda_j^2}, 
\label{eq:euler_k_ext_sol}
\end{align}
which involves a truncation error of $O(h^3)$, asymptotically smaller than $O(h^2)$.

In the sampling process, we set the initial condition at the next denoising step, $t_{i-k}$, as $\tilde\x_{t_{i-1}}^{(k)}$, and repeatedly perform the proposed extrapolation technique every $k$ steps.
Because this approach provides provably more accurate solutions at every $k$ steps, we can reduce error propagation and expect better quality of generated examples.

The proposed method is applicable to first-order methods in general, including DDIM~\citep{song2021denoising}, which is arguably the most widely used DPM sampler. 
In this context, we bring the interpretation of DDIM as the Euler method applied to the following ODE:
\begin{align}
    d\y = \epsilon_\theta(\x(t), t)d\gamma,    
\end{align}
where $\y(t)=\x(t)\sqrt{1+\gamma(t)^2}$ and $\gamma(t) = \sqrt{\frac{1-\alpha_t^2}{\alpha_t^2}}$ in the variance-preserving diffusion process~\citep{song2021scorebased}, \ie, $\bm{p}_t(\x|\x_0) = \mathcal{N}(\alpha_t \x_0, (1-\alpha_t^2) \I)$.
Thus, instead of using the time grid, we compute $\lambda_j$'s in \cref{eq:euler_k_ext_sol} in terms of $\gamma(t)$'s, replacing $t$ with the corresponding $\gamma(t)$ in the computation, while the other procedures remain unchanged.

RX-Euler (RX-DDIM) does not require additional NFEs beyond the number of time steps, as the first prediction of every $k$-step-interval can be stored during the computation of $\hat\x^{(k)}$ and reused to obtain $\hat\x^{(1)}$. 
The only extra computation involves a linear combination of two estimates, which is negligible compared to the forward evaluations of DPMs.

\subsection{RX-DPM with higher-order solvers}
\label{subsec:rx-ode}
We now present the algorithm for general ODE samplers of DPMs including high-order solvers.  
When the extrapolation occurs every $k$ steps, as in \cref{subsec:rx-euler}, the error of $\hat \x^{(1)}_{t_{i-k}}$, resulting from a single iteration of an arbitrary ODE method, is given by 
\begin{align}
    \hat \x^{(1)}_{t_{i-k}} = \x^*_{t_{i-k}} + \Vc h^p + O(h^q)
    \label{eq:ode_e1}
\end{align}
for $0<p<q$ and $\Vc \neq \bm{0}$.
For $\hat \x^{(k)}_{t_{i-k}}$, we extend the form of the linear error accumulation observed in \cref{eq:euler2} to obtain the following equation:
\begin{align}
    \hat \x^{(k)}_{t_{i-k}} = \x^*_{t_{i-k}} + \Vc\sum_{j=1}^k \lambda_j^p  h^p  + O(h^q).
    \label{eq:ode_e2}
\end{align}
Note that \cref{eq:ode_e2} does not hold in general; however, we consider this simplified assumption reasonable, as it is consistent with the standard assumption of Richardson extrapolation under uniform discretization (see \cref{app:justification}).
Finally, by solving the linear system of \cref{eq:ode_e1,eq:ode_e2}, the extrapolated solution is given by
\begin{align}
    \tilde{\x}_{t_{i-k}}^{(k)} = \frac{\hat \x^{(k)}_{t_{i-k}} - \sum_{j=1}^k \lambda_j^p \hat \x^{(1)}_{t_{i-k}}}{1-\sum_{j=1}^k \lambda_j^p}, 
    \label{eq:ode_k_ext_sol}
\end{align}
which approximates $\x_{t_{i-k}}^*$ with a truncation error of $O(h^q)$, asymptotically smaller than $O(h^p)$.

A limitation of this approach is that estimating $\hat \x^{(1)}_{t_{i-k}}$ and $\hat \x_{t_{i-k}}^{(k)}$ through na\"{i}ve applications of higher-order solvers requires additional network evaluations compared to baseline sampling.
However, since $\hat \x^{(1)}_{t_{i-k}}$ is accessible from the network predictions made for the computation of $\hat \x_{t_{i-k}}^{(k)}$, applying our method to higher-order solvers does not increase the NFE provided that intermediate predictions are properly stored.
We next discuss how this is achieved using specific examples of high-order ODE solvers; the generalization to other solvers is mostly straightforward.

Before moving forward, we note that high-order solvers typically rely on interpolation-based techniques, such as the Runge-Kutta method~\citep{suli2003introduction} and linear multistep method~\citep{timothy2017numerical}, where the former employs evaluations at multiple intermediate points, while the latter leverages evaluations from previous steps. 

\vspace{-1mm}
\paragraph{RX-Runge-Kutta}
We consider applying our method with $k = 2$ to the second-order Runge-Kutta method.
A sequence of one-step estimates are given by 
\begin{align}
\hat \x_{t_{i-1}}^{(1)} &= \x_{t_{i}} - (t_{i} - t_{i-1})(a_1 \z_{i} + a_2 \z_{i-\delta}) ~~\text{and} \\
\hat \x_{t_{i-2}}^{(2)} & = \hat{\x}_{t_{i-1}}^{(1)} - (t_{i-1} - t_{i-2})(a_1 \z_{i-1} + a_2 \z_{i-1-\delta}).
\end{align}
where $\z_{j} = \epsilon_\theta(\x(t_{j}), t_{j})$ for $t_{j-1} < t_{j-\delta} \leq t_{j}$. 
Then, we can express the single combined-step estimate at $t_{i-2}$ as 
\begin{equation}
\hat \x_{t_{i-2}}^{(1)} = \x_{t_{i}} - (t_{i} - t_{i-2})(a_1 \z_{i} + a_2 \z_{i-\delta'}).
\end{equation}
Since $\z_{i}$ is reusable after the calculation of $\hat \x_{t_{i-1}}^{(1)}$, we only need to compute $\z_{i-\delta'}$, which is approximated as $\z_{i-1}$ or $\z_{i-1-\delta}$, depending on the proximity of its time step.
This approach allows us to efficiently extrapolate the solutions without compromising the quality of the generated samples.

\vspace{-1mm}
\paragraph{RX-Adams-Bashforth}
Suppose that, by the $s$-step Adams-Bashforth method, extrapolation is performed on a grid with an interval of $h$ every $k$ steps.
For predefined $b_j$'s, we are given
\begin{align}
\hat\x_{t_{i-k}}^{(k)} &= \hat\x_{t_{i-k+1}} + h\sum_{j=0}^{s} b_j\epsilon_\theta(\hat\x_{t_{i-k+j}}, t_{i-k+j}).
\label{eq:adam_bashforth}
\end{align}
Then, we compute $\hat\x_{t_{i-k}}^{(1)}$ with a step size of $kh$ for extrapolation as
\begin{align}
\hat\x_{t_{i-k}}^{(1)} &= \hat\x_{t_{i}} + kh\sum_{j=0}^{s} b_j\epsilon_\theta(\hat\x_{t_{i-k+jk}}, t_{i-k+jk})
\label{eq:adam_bashforth_extrapolation}
\end{align}
which requires no additional NFEs by storing the previous network evaluations.

\cref{alg:rx_algorithm} summarizes the procedure of the proposed method with a generic ODE solver under the assumption that $N$ is a multitude of $k$ for simplicity; it is simple to handle the last few steps by either adjusting $k$ for the remaining steps or skipping the extrapolation.

\begin{algorithm}[t]
    \caption{Sampling of RX-DPM}
    \label{alg:rx_algorithm}
    \begin{algorithmic}[1]
        
        \Require $\epsilon_\theta(\cdot)$, $N$, $T = t_N > \ldots > t_0 = 0$
        \State \textbf{Input:} $k$, $\Phi(\cdot)$~(ODE solver), $p$
        
        \State $\x_T \sim \bm{p}_T(\x)$

        \For{$i=1$~\textbf{to}~$N$}
            \vspace{1mm} 
            \State {\textit{\# Initialization}}
            \If{$i~\text{mod}~k =1$}
                \State $h \leftarrow t_{N-i+1} - t_{N-i-k+1}$
                \State $\hat\x_{t_{N-i+1}}^{(k)} \leftarrow \x_{t_{N-i+1}}$
            \EndIf

            \vspace{1mm}
            \State $\lambda_i \leftarrow (t_{N-i+1} - t_{N-i})/h$ 
            \State $\hat\x_{t_{N-i}}^{(k)} \leftarrow \Phi(\hat\x_{t_{N-i+1}}^{(k)},t_{N-i+1}, t_{N-i};\epsilon_\theta(\cdot))$ \Comment{Store $\epsilon_\theta(t)$'s if neccessary.}
            
            \vspace{2mm} 
            \State {\textit{\# Extrapolation~(\cref{eq:ode_k_ext_sol})} }
            \If{$i~\text{mod}~k = 0$}
                \State $\hat\x_{t_{N-i}}^{(1)} \leftarrow \Phi(\x_{t_{N-i}+h},t_{N-i}+h, t_{N-i};\epsilon)$ \Comment{No NFE required.}
                \State $\tilde\x_{t_{N-i}}^{(k)} \leftarrow \frac{\hat\x^{(k)}_{t_{N-i}}-\sum_{j=i}^{i+k-1} \lambda_j^p \hat\x^{(1)}_{t_{N-i}}}{1-\sum_{j=i}^{i+k-1} \lambda_j^p}$ 
                \State $\x_{t_{N-i}} \leftarrow \tilde\x_{t_{N-i}}^{(k)}$ 
            \EndIf
        
        \EndFor
        \State \textbf{return} $\x_{t_0}$
    \end{algorithmic}
\end{algorithm}

\subsection{Analysis on global truncation errors}
We perform an analysis on global truncation errors of the Euler method and RX-Euler under the same NFEs.
Assume that we solve an ODE satisfying Lipschitz condition from $t=1$ to $t=0$ with $\text{NFEs} = N$.
\vspace{-2mm}
\paragraph{Euler}
\label{subsubsec:euler}
Since the Euler method requires a single network evaluation for each time step, the number of time steps allowed is $N$.
Since the local truncation error of the Euler method on step size of $h=1/N$ is expressed as $ch^2 + O(h^3)$, the global truncation error is given by
\begin{align}
    (ch^2 + O(h^3)) \cdot N = \frac{c}{N} + O(N^{-2}) = O(N^{-1}).
\end{align}

\vspace{-2mm}
\paragraph{RX-Euler}
If RX-Euler performs extrapolation every $k$ steps, the extrapolation happens $N/k$ times, where $N$ is equal to the NFEs for the Euler method.
The local truncation error for RX-Euler over each $k$ steps, which has the interval of $h=k/N$, is expressed as $c'h^3 + O(h^4)$ and therefore the global truncation error is given by
\begin{align}
    (c'h^3 + O(h^4)) \cdot \frac{N}{k} = \frac{k^2c'}{N^2} + O(N^{-3}) = O(N^{-2}).
\end{align}
RX-Euler exhibits a higher convergence rate of the global truncation error compared to the Euler method by one order of magnitude.
Using the same approach, we can also demonstrate that the proposed method achieves faster convergence of the global truncation error for higher-order solvers.


\section{Experiment}
\label{sec:experiment}
\subsection{Implementation details}
\label{subsec:implement}
We conduct the experiment with EDM~\citep{karras2022elucidating}, Stable Diffusion V2\footnote{\url{https://github.com/Stability-AI/stablediffusion}, \text{v2-1\_512-ema-pruned.ckpt}}~\citep{rombach2022high}, DPM-Solver~\citep{lu2022fastode}, and PNDM~\citep{liu2022pseudo} using their official implementations and provided pretrained models.
Throughout all experiments, we retain the default settings from the official codebases, except for additional hyperparameters related to the proposed method.  
For experiments with EDM, DPM-Solver, and PNDM as backbones, we generate 50K images and compute FID~\citep{heusel2017gans} using the evaluation code provided in their implementations.  
To evaluate Stable Diffusion V2 results, we use the PyTorch implementation for the computation of FID\footnote{\url{https://github.com/mseitzer/pytorch-fid}} and CLIP score\footnote{\url{https://huggingface.co/openai/clip-vit-base-patch32}} with the patch size of $32\times32$.

\subsection{Validity test}
\begin{wrapfigure}{r}{0.5\textwidth}
    \centering
    \setlength{\tabcolsep}{0.2mm}
    \scalebox{1.0}{
        \begin{tabular}{cc}
        \includegraphics[width=0.48\linewidth]{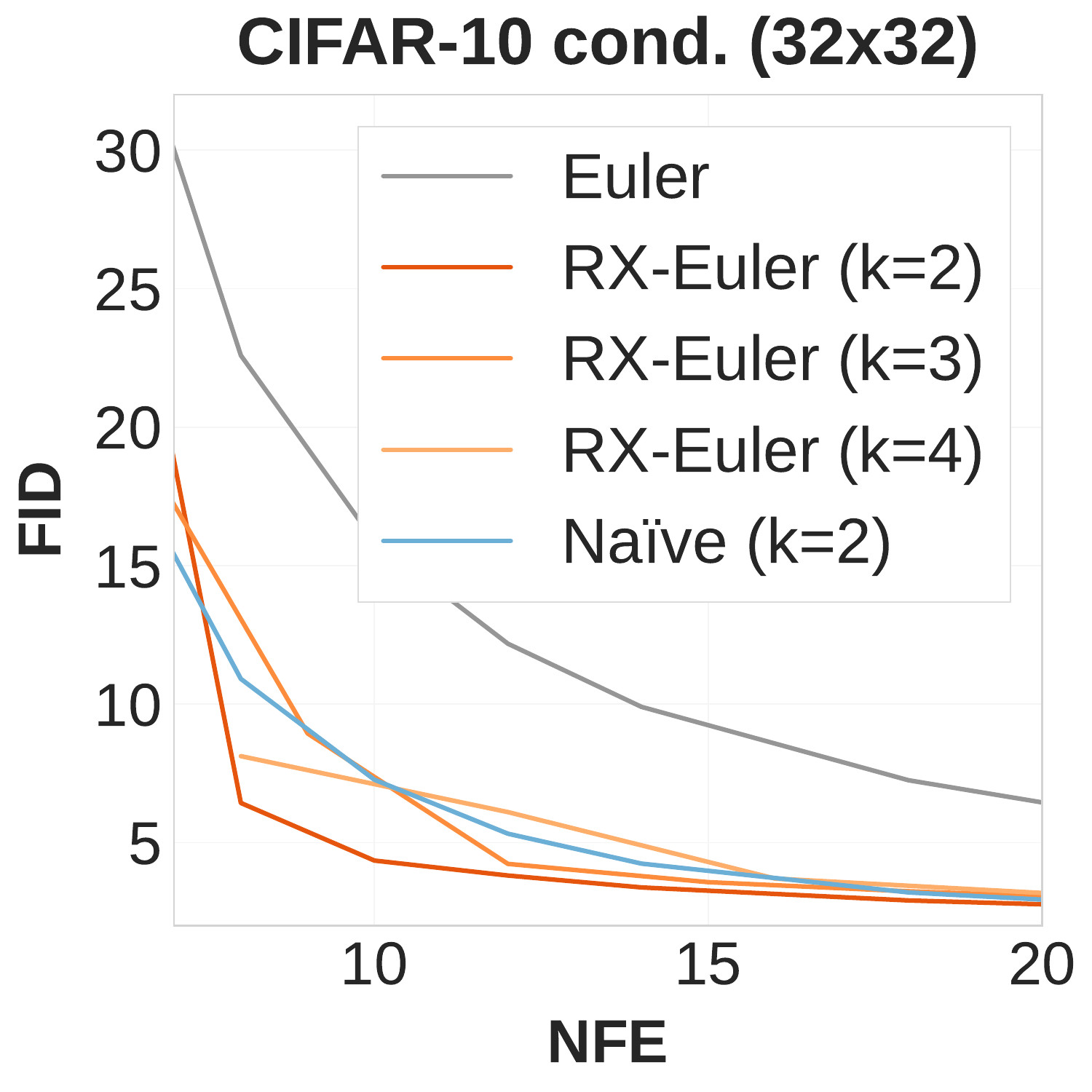} &
        \includegraphics[width=0.48\linewidth]{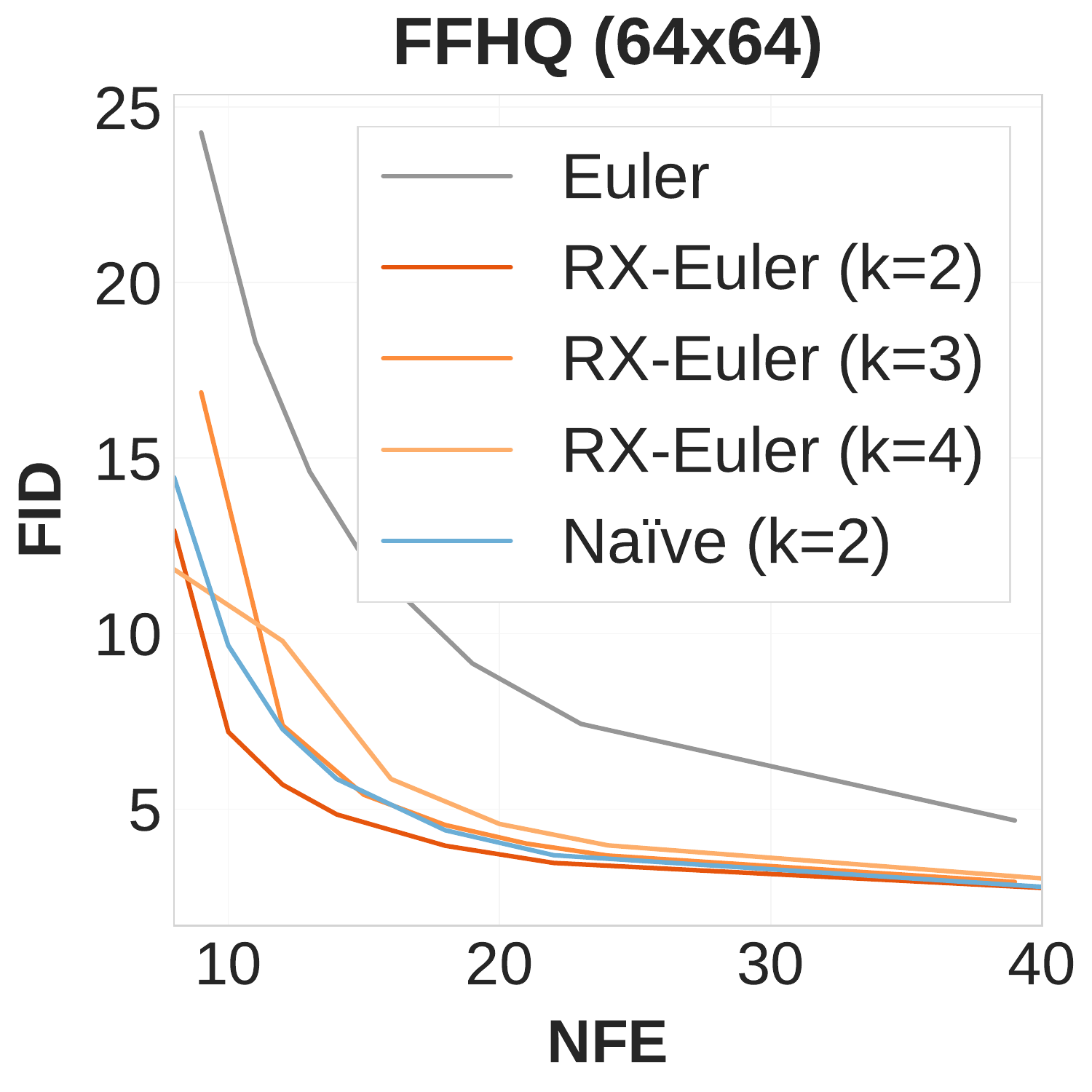}
        \end{tabular}
    }
    \vspace{-2mm}
    \caption{
    Effect of extrapolation on the Euler method with different $k$'s. 
    }
    \label{fig:k}
\end{wrapfigure}

We first evaluate the effectiveness of RX-Euler under the EDM backbone for $k\in\{2,3,4\}$, where smaller $k$ values correspond to more frequent extrapolation over the same number of time steps.
As shown in \cref{fig:k}, RX-Euler consistently achieves significantly better FID scores than the Euler method across all values of $k$.
In particular, when extrapolation is applied every two steps, \ie, $k=2$,  our approach achieves the best performance across a wide range of NFEs.
This indicates that more frequent extrapolation leads to more accurate intermediate predictions, effectively mitigating error accumulation in the final samples.
Meanwhile, the curves for $k \geq 3$ remain closer to that of $k=2$ than to the Euler method, 
empirically validating the reduced truncation error derived in \cref{eq:euler_k_ext_sol} for general $k$.
In other words, even sparse extrapolation still has a significant impact on the output quality.
For the rest of our results, we set $k=2$.

\label{subsec:effective_rx}
We also compare the proposed method with conventional Richardson extrapolation, as formulated in \cref{eq:ric_extra_sol} with $k=2$, which employs fixed coefficients.  
This baseline is labeled as Na\"{i}ve ($k=2$) in \cref{fig:k}.  
When comparing RX-Euler ($k=2$) with Na\"{i}ve ($k=2$), we find that our method produces superior results.  
This implies that extrapolation coefficients adapted to arbitrary time step scheduling are more effective than fixed coefficients derived from uniformly discretized time steps.  
In other words, the proposed grid-aware coefficients enable more effective extrapolation than deterministic ones, better capturing the varying importance of precision in DPMs over time~\citep{karras2022elucidating}.

\vspace{-1mm}
\subsection{Quantitative comparisons on EDM backbone}
\label{subsec:edm}
We compare RX-Euler with other methods on four different datasets---CIFAR-10~\citep{krizhevsky2009learning}, FFHQ~\citep{karras2019style}, AFHQv2~\citep{choi2020stargan}, and ImageNet~\citep{deng2009imagenet}---using the EDM~\citep{karras2022elucidating} backbone.
Following standard practice, we evaluate the performance of class-conditional generation on CIFAR-10 and ImageNet while testing unconditional generation on the other two datasets.
In this experiment, we include Heun's method, LA-DPM~\citep{zhang2023lookahead}, and IIA~\citep{zhang2024iia} for comparisons with our approach, RX-Euler.
Note that both Heun's method and RX-Euler are second-order numerical solvers, offering higher accuracy than the Euler method, while LA-DPM and IIA are techniques refining baseline sampling.
To reproduce the results of LA-DPM, we use the Euler method, as it achieves better performance than Heun's method.
For IIA~\citep{zhang2024iia}, we present results from the better-performing variant, selected between IIA and BIIA, as indicated in the original paper.

\begin{figure}[t]
    \setlength{\tabcolsep}{0mm}
    \begin{tabular}{cccc}
        \includegraphics[width=0.25\textwidth]{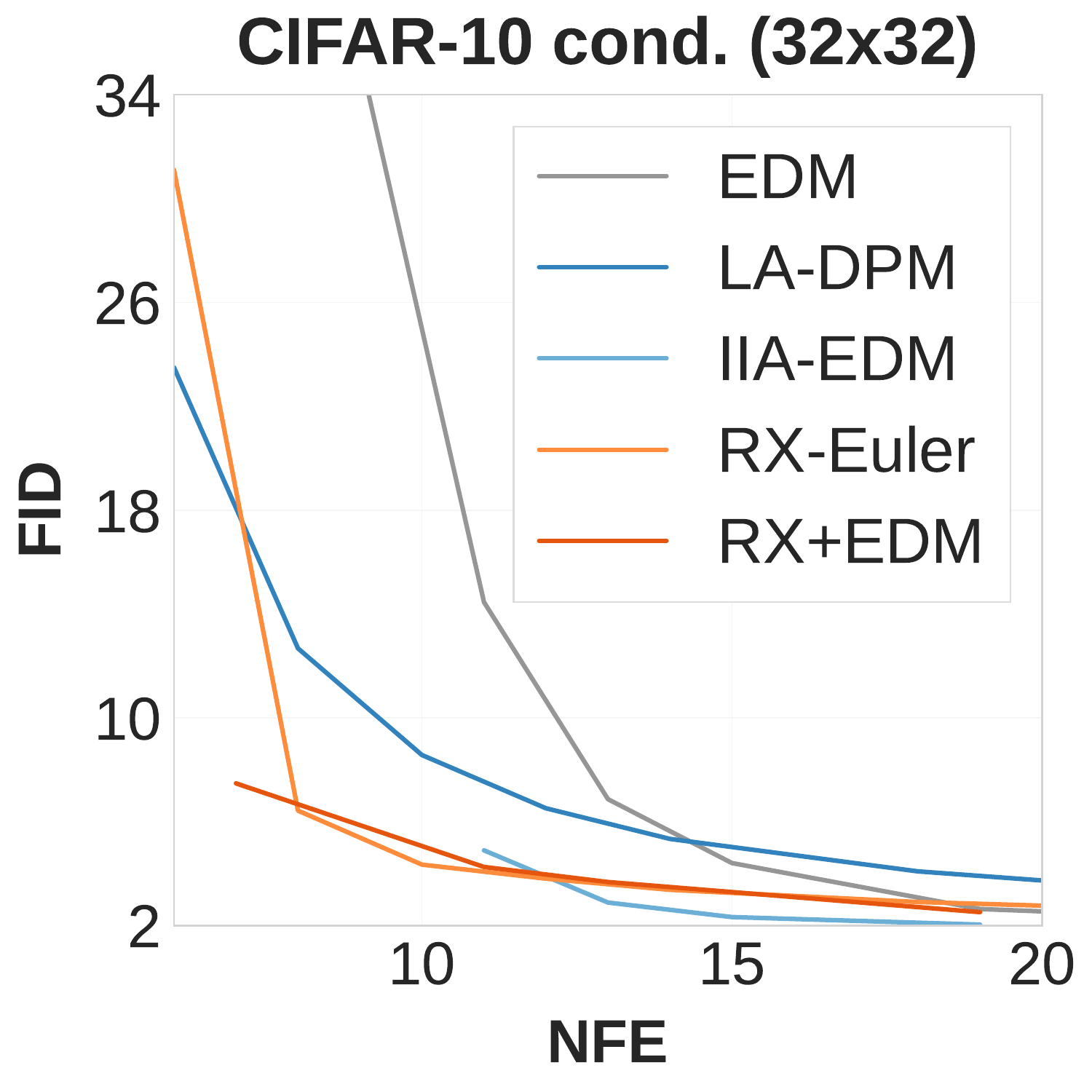}
        &
        \includegraphics[width=0.25\textwidth]{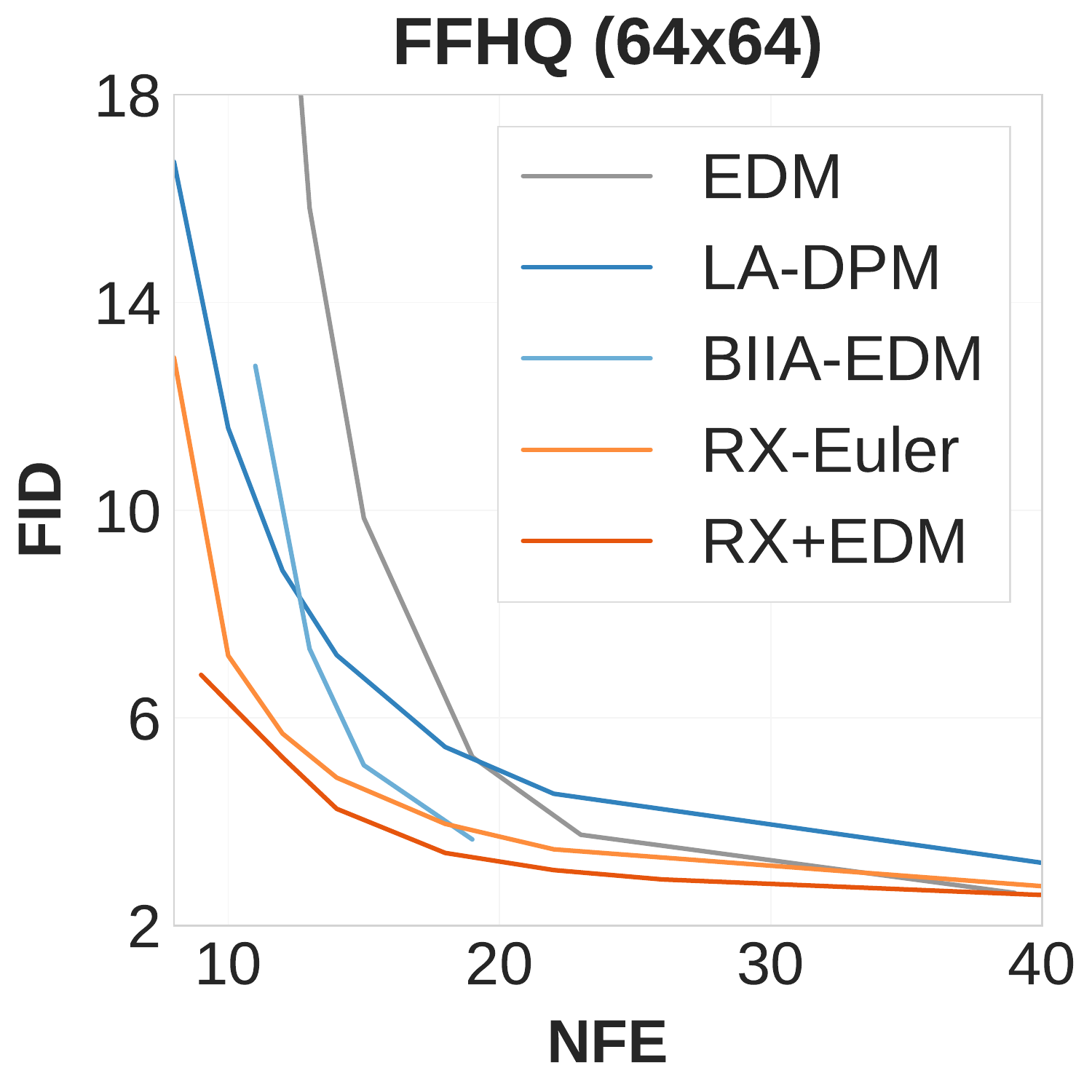}
        &
        \includegraphics[width=0.25\textwidth]{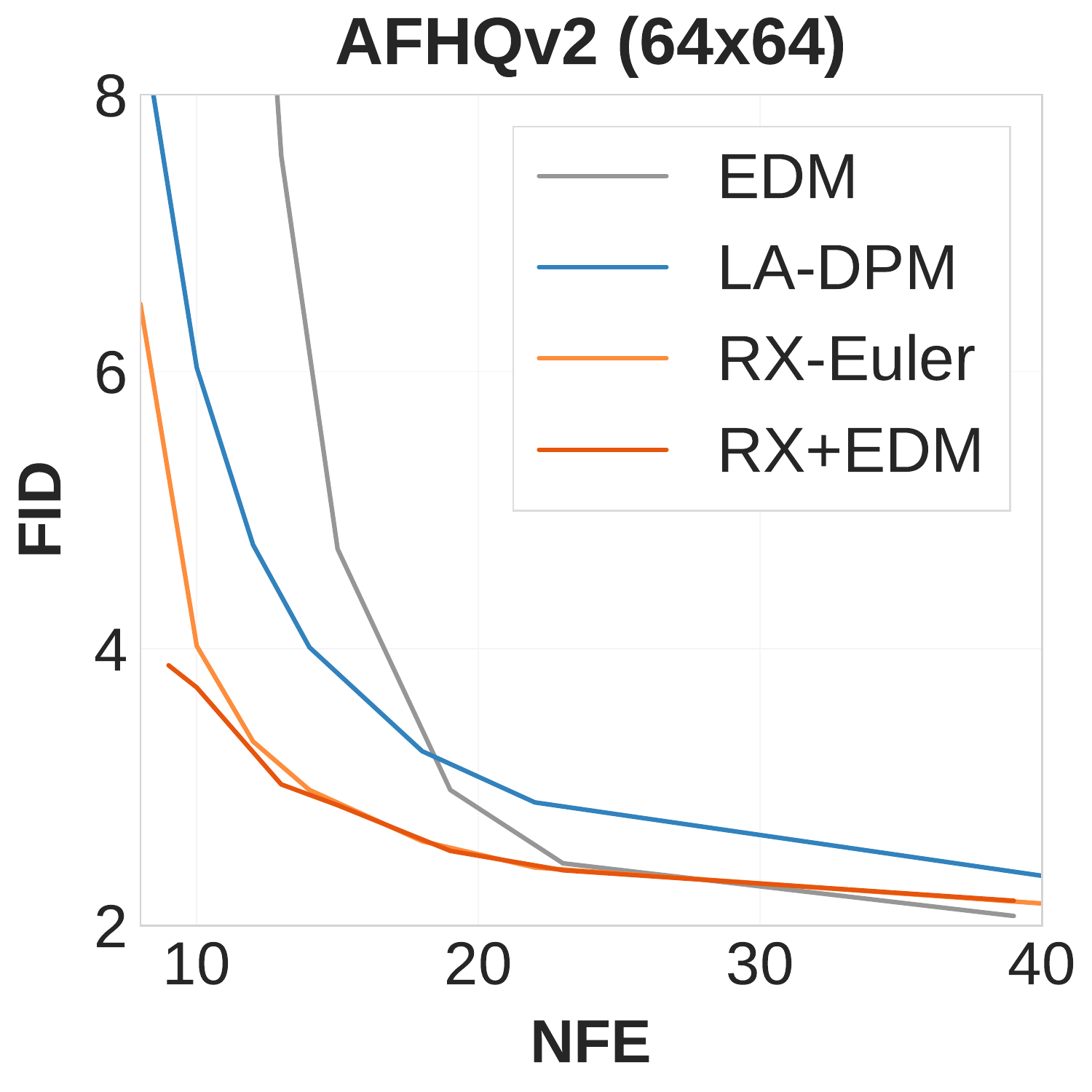} 
        &
        \includegraphics[width=0.25\textwidth]{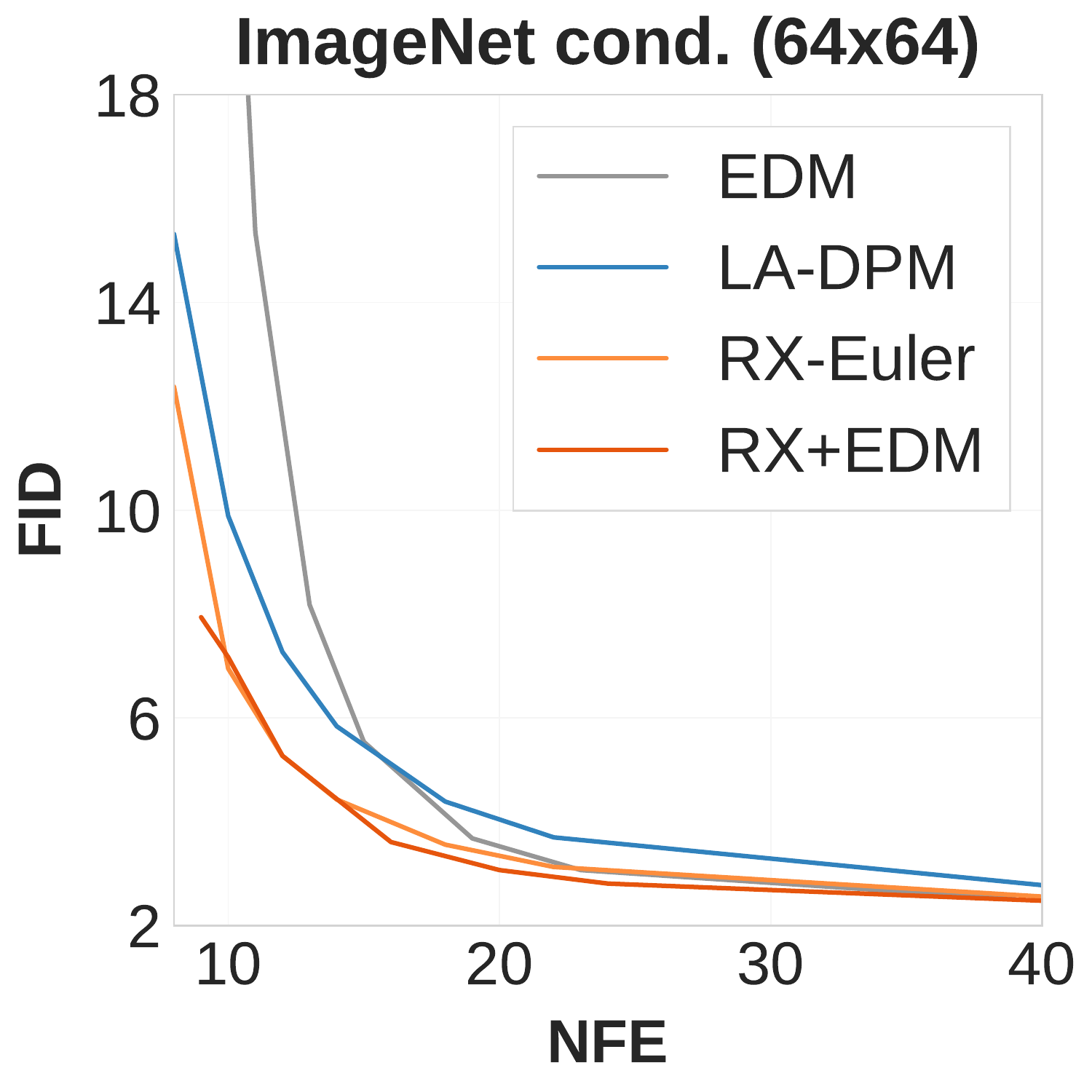} \\ 
    \end{tabular}
    \vspace{-3mm}
    \caption{FIDs of RX-Euler, Heun's method (labeled as EDM), LA-DPM and IIA (or BIIA) by varying NFEs on the CIFAR-10, FFHQ, AFHQv2, and ImageNet datasets using the EDM backbone.}
    \label{fig:rx_edm}
\end{figure}

\cref{fig:rx_edm} presents a comparison of FID scores across the evaluated methods over a broad range of NFEs.
RX-Euler consistently outperforms the other approaches, particularly at lower NFEs, demonstrating its effectiveness in fast sampling scenarios.
While it occasionally falls behind LA-DPM or IIA on CIFAR-10 within specific intervals, its overall performance remains more stable and superior across the other three datasets, which pose greater challenges due to higher resolutions and increased data diversity.

In the comparison between RX-Euler and Heun's method, while RX-Euler excels at lower NFEs, Heun's method performs slightly better at larger NFEs. 
This implies that selecting a more appropriate solver for each interval---between RX-Euler (extrapolation) and Heun's method (interpolation)---could yield better results. 
In this regard, one might expect that in the early steps---where predictions are closer to noise and thus less accurate---interpolation tends to be more stable than extrapolation.
Based on this reasoning, we experiment with a hybrid approach of RX-Euler and Heun's method, labeled as RX+EDM in \cref{fig:rx_edm}.
We find strong performance of this approach when RX-Euler is selectively applied to the middle or last (low-noise) few steps, outperforming both Heun's method and RX-Euler. 
This indicates that there is still room for improvement in our algorithm and provides another direction for future work.

\begin{figure}[t]
    \centering
    \renewcommand{\arraystretch}{0.7}
    \scalebox{0.98}{
    \setlength{\tabcolsep}{1pt} \hspace{-0.5mm}
        \hspace{-3mm}
    \begin{tabular}{ccccc}
    & $\text{NFEs}=15$ & $\text{NFEs}=50$ & $\text{NFEs}=15$ & $\text{NFEs}=50$ \\ 
    \rotatebox{90}{\hspace{12mm}{DDIM}}~ &
    \includegraphics[width=0.24\textwidth]{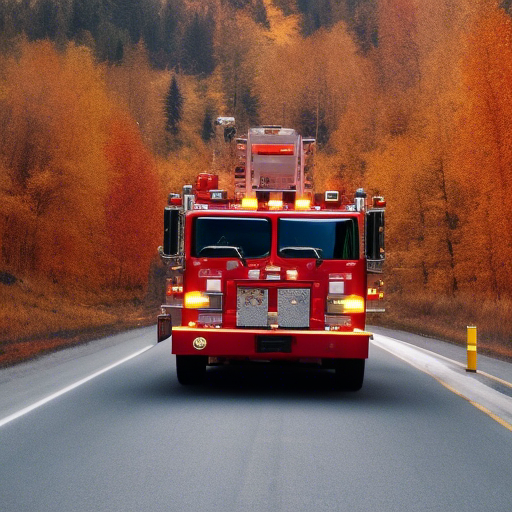} &\includegraphics[width=0.24\textwidth]{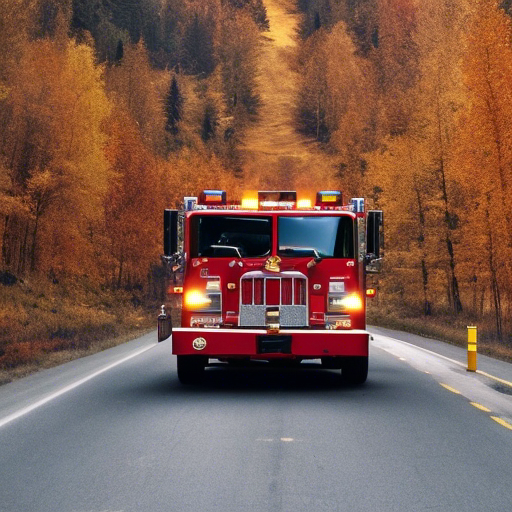} &
    ~\includegraphics[width=0.24\textwidth]{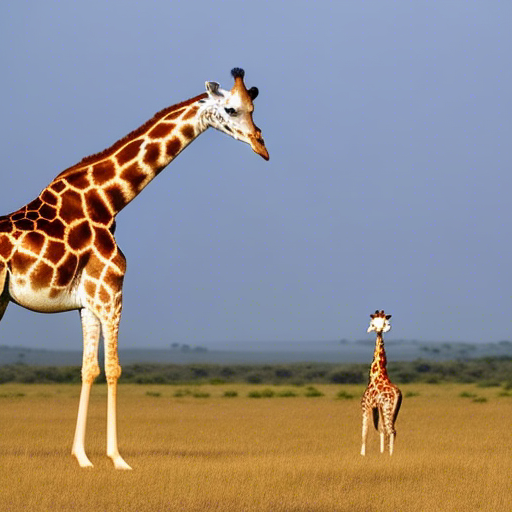} &\includegraphics[width=0.24\textwidth]{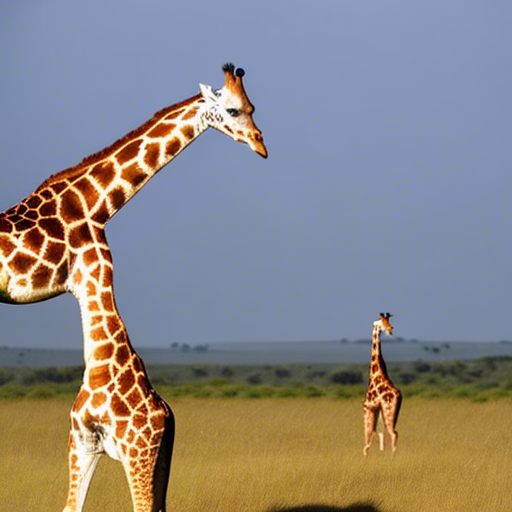} \\
    \rotatebox{90}{\hspace{9mm}{RX-DDIM}}~ & 
    \includegraphics[width=0.24\textwidth]{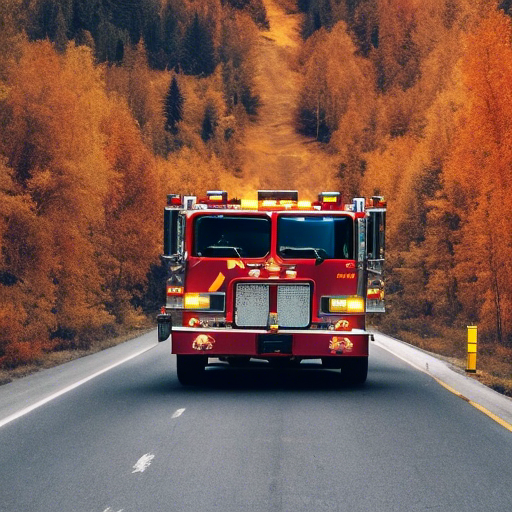} &\includegraphics[width=0.24\textwidth]{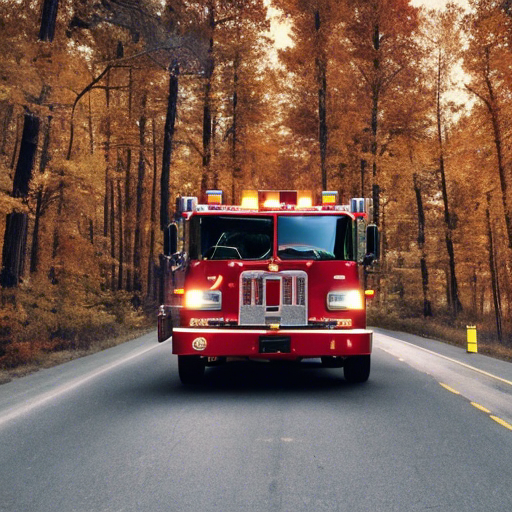} &
    ~\includegraphics[width=0.24\textwidth]{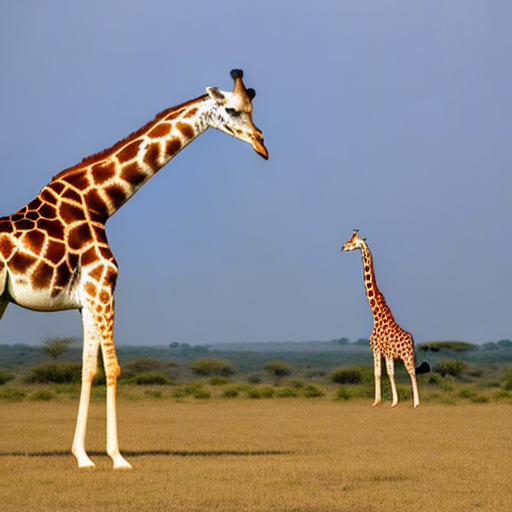} &\includegraphics[width=0.24\textwidth]{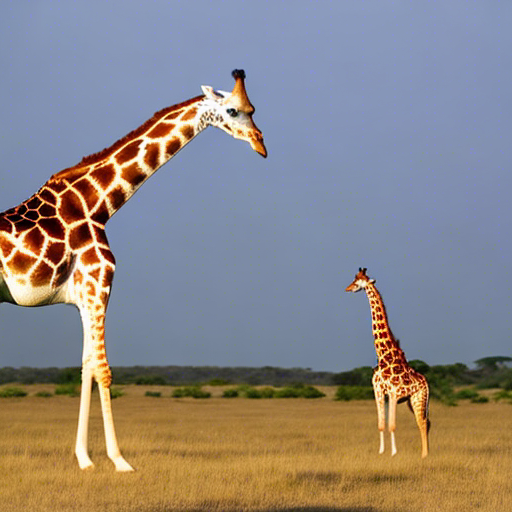} \\
    &\multicolumn{2}{c}{\textit{``A red fire truck driving down a road.''}}
    &\multicolumn{2}{c}{\textit{``A pair of giraffe standing in a big open area.''}}
    \end{tabular}
    }
    \vspace{-1mm}
    \caption{Qualitative results of DDIM and RX-DDIM based on Stable Diffusion V2. 
    RX-DDIM produces sharper and more detailed backgrounds, especially evident in the left example.
    On the right, RX-DDIM generates more realistic giraffes, whereas DDIM struggles at  $\text{NFEs}=50$, failing to properly render the giraffe.}
    \label{fig:rx_sd_main}
    \vspace{-2mm}
\end{figure}
\begin{table}[H]
    \centering
    \caption{FID and CLIP scores of DDIM and RX-DDIM using Stable Diffusion V2.}
    \vspace{-3mm}
    \label{tab:rx_sd}
    \renewcommand{\arraystretch}{1.0}
    \setlength{\tabcolsep}{1.8mm}
    \scalebox{0.8}{
        \begin{tabular}{lccccccccccccc}
            \toprule
            NFEs & & & \multicolumn{2}{c}{15} & & \multicolumn{2}{c}{20} & & \multicolumn{2}{c}{30} & & \multicolumn{2}{c}{50} \\
            \cmidrule{1-2} \cmidrule{4-5} \cmidrule{7-8} \cmidrule{10-11} \cmidrule{13-14}

            Method ~~\textbackslash~~ Metric &  &  & FID~($\downarrow$) & CLIP~($\uparrow$) &  & FID~($\downarrow$) & CLIP~($\uparrow$) &  & FID~($\downarrow$) & CLIP~($\uparrow$)&  & FID~($\downarrow$) & CLIP~($\uparrow$) \\
            \cmidrule{1-2} \cmidrule{4-5} \cmidrule{7-8} \cmidrule{10-11} \cmidrule{13-14}

            DDIM &  &  & 19.15 & \B 31.727&&18.43&	31.716&&19.00& 31.750&&18.65&	31.711 \\
            RX-DDIM &  &  & \B 17.24&	31.629&&\B 17.12&	\B 31.721&&\B 17.62&\B 31.781&&\B 17.83&\B 31.727 \\
            \bottomrule
        \end{tabular}
    }
    \vspace{-1mm}
\end{table}

\subsection{Comparisons on Stable Diffusion}
\label{subsec:sd}
We apply RX-DDIM to Stable Diffusion V2, which provides various conditional generations. 
For evaluation, we generate 10K $512\times512$ images from unique text prompts in the COCO2014~\citep{cocodataset} validation set and compute FID and CLIP scores on resized $256\times256$ images.
As shown in \cref{tab:rx_sd}, our method also performs well on large models.
However, we observe that RX-DDIM yields lower CLIP scores at $\text{NFEs}=15$, which we attribute to classifier-guidance scales.
According to \cite{rombach2022high}, optimal classifier-free guidance scales differ across models.
Since the default setting is tuned for DDIM, RX-DDIM may benefit from further optimization.
\cref{fig:rx_sd_main} presents qualitative comparisons between DDIM and RX-DDIM, highlighting the superior image quality of RX-DDIM.
Notably, RX-DDIM generates images with more vivid colors, sharper textures, and more realistic object depictions, leading to an overall more natural appearance.
We provide more examples for qualitative comparisons between DDIM and RX-DDIM in \cref{fig:rx_sd1,fig:rx_sd2} of \cref{app:qual}.

\begin{table}[t]
    \centering
    \setlength{\tabcolsep}{3mm}
    \caption{FID scores of DPM-Solvers~\citep{lu2022fastode} and RX-DPMs applied to DPM-Solvers on CIFAR-10 and LSUN Bedroom datasets. All baseline results are reproduced under the same setting as RX-DPMs. }
    \vspace{-1mm}
    \scalebox{0.8}{
    \begin{tabular}{lccccccccc}
        \toprule
        & \multicolumn{4}{c}{CIFAR-10 (32$\times$32)} &  & \multicolumn{4}{c}{LSUN Bedroom (256$\times$256)} \\
        \cmidrule{2-5} \cmidrule{7-10}
        Method ~~\textbackslash~~ NFEs & 9 & 10 & 12 & 15 & & 9 & 10 & 12 & 15 \\
        \midrule
        DPM-Solver-2 &-- & 15.06 & 11.33 & 7.36 & &-- & 14.67 & 11.38 & 6.44 \\
        RX-DPM-Solver-2 &-- & \B12.94 & \ \ \B9.80 & \B6.53 & &-- & \B12.66 & \B10.13 & \B5.72 \\
        \midrule
        DPM-Solver-3 & 12.39 & --& 6.76 & 5.00 & & 8.79 &-- & 5.37 & \B4.04 \\
        RX-DPM-Solver-3 & \B11.50 &-- & \B6.62 & \B4.85 & & \B8.12 &-- & \B5.18 & \B4.04 \\
        \bottomrule
    \end{tabular}
    }
    \label{tab:dpm_solver}
\end{table}
\begin{table}[t]
    \centering
    \setlength{\tabcolsep}{2mm}
    \caption{FID scores of two types of PNDM~\citep{liu2022pseudo} and RX-DPMs applied to each solver on CIFAR-10, CelebA and LSUN Church datasets. Note that S-PNDM and F-PNDM require 1 and 9 additional NFEs to the number of time steps, respectively. The baseline results are copied from PNDM~\citep{liu2022pseudo}.}
    \vspace{-1mm}
    \scalebox{0.8}{
    \begin{tabular}{lccccccccccc}
        \toprule
        & \multicolumn{3}{c}{CIFAR-10 (32$\times$32)}
        & & \multicolumn{3}{c}{CelebA (64$\times$64)} 
        & & \multicolumn{3}{c}{LSUN Church (256$\times$256)} \\
        \cmidrule{2-4} \cmidrule{6-8} \cmidrule{10-12}
        Method ~~\textbackslash~~ \# of steps
        & 5 & 10 & 20 & & 5 & 10 & 20 & & 5 & 10 & 20 \\
        \midrule
        S-PNDM 
        & \B18.3 \ \ &	8.64 & 5.77 & &15.2 \ \ & 12.2 \ \ \ \  & 9.45 & & \B20.5 \ \ & 11.8 \ \ & 9.20 \\
        RX-S-PNDM & 19.69 & \B7.64 & \B4.72 & & \B11.56 & \B9.22 & \B6.89 & & 21.15 & \B10.96 & \B8.96 \\
        \midrule
        F-PNDM 
        & 18.2 \ \ & 7.05 & 4.61 & & 11.3 & 7.71 & 5.51 & & 14.8 & \B8.69 & \B9.13 \\
        RX-F-PNDM & -- & \B6.60 & \B3.99 & &-- & \B7.10 & \B4.99 & & -- & 8.85 & 9.41 \\
        \bottomrule
    \end{tabular}
    }
    \label{tab:pndm}
\end{table}
\subsection{Comparisons on higher-order solvers}
\label{subsec:high_order}
We further apply the proposed method to advanced ODE samplers with higher-order accuracy.
\cref{tab:dpm_solver} presents the effectiveness of RX-DPM when applied to DPM-Solvers~\citep{lu2022fastode} on CIFAR-10 and LSUN Bedroom~\citep{yu2015lsun}.
Among the variations of DPM solvers, we utilize the single-step versions of DPM-Solver-2 and DPM-Solver-3. 
Note that, since a single-step DPM-solver-$n$ can be considered as an $n^{\text{th}}$-order Runge-Kutta-like solver, we apply RX-DPM with $p=n+1$ in \cref{eq:ode_k_ext_sol} for the DPM-solver-$n$. 
Additionally, we compare our method with another accelerated diffusion sampler, DEIS~\citep{zhang2023fast}, for class-conditioned image generation on ImageNet ($64\times64$) in \cref{tab:deis} of \cref{app:deis}.
As shown in the results, RX-DPM consistently achieves the best performance across all NFEs.

As another type of advanced sampler, we consider PNDM~\citep{liu2022pseudo}.
S-PNDM and F-PNDM employ linear multistep methods, specifically the 2-step and 4-step Adams-Bashforth methods, respectively, except for the initial few time steps.
Accordingly, we apply RX-DPM with $p=3$ and $p=5$ in \cref{eq:ode_k_ext_sol} for S-PNDM and F-PNDM, respectively.
The results on the CIFAR-10, CelebA~\citep{liu2015faceattributes}, and LSUN Church~\citep{yu2015lsun} datasets are presented in \cref{tab:pndm}.
While RX-DPM improves performance in most cases, an exception arises with F-PNDM on the LSUN Church dataset, where RX-DPM does not provide a clear advantage.
Upon analysis, we observe that the baseline performance of F-PNDM is highest when using 10 time steps and degrades as the number of steps increases~\citep{liu2022pseudo}.
Since RX-DPM enhances accuracy by leveraging improvements in the baseline solver with finer time steps, its effectiveness is limited when the baseline itself deteriorates under finer discretization.
A similar phenomenon with F-PNDM on LSUN datasets has also been reported in IIA~\citep{zhang2024iia}.


\vspace{-1mm}

\section{Conclusion}
\label{sec:conclusion}
\vspace{-1mm}
We introduced RX-DPM, an advanced ODE sampling method for DPMs that leverages extrapolation based on two ODE solutions derived from different discretizations of the same time interval. 
Our algorithm computes the optimal coefficients for arbitrary time step scheduling without additional training and incurs no extra NFEs by utilizing past predictions.
This approach effectively reduces truncation errors, resulting in improved sample quality.
Extensive experiments on well-established baseline models and datasets confirm that RX-DPM surpasses existing sampling methods, offering a more efficient and accurate solution for DPMs.

\subsubsection*{Acknowledgements}
\label{subsec:acknowledgement}
This work was partly supported by Samsung Research, Samsung Electronics Co., Ltd., and the Institute of Information \& communications Technology Planning \& Evaluation (IITP) grants [RS-2022-II220959 (No.2022-0-00959), (Part 2) Few-Shot Learning of Causal Inference in Vision and Language for Decision Making; No.RS-2021-II212068, AI Innovation Hub (AI Institute, Seoul National University); No.RS-2021-II211343, Artificial Intelligence Graduate School Program (Seoul National University)] funded by the Korea government (MSIT).

\subsubsection*{Ethics statement}
\label{subsec:ethics}
This paper utilizes pretrained diffusion probabilistic models to generate high-quality images while prioritizing efficiency. 
As a result, our approach does not inherently introduce hazardous elements. 
Nonetheless, we acknowledge the potential for unintended misuse, including the synthesis of inappropriate or sensitive content.

\subsubsection*{Reproducibility statement}
\label{subsec:reproducibility}
We provide detailed implementation instructions and reproducibility guidelines in \cref{subsec:implement}, along with the pseudo-code of the core algorithm in \cref{alg:rx_algorithm}. 
The full implementation is available at \url{https://github.com/jin01020/rx-dpm}.

\bibliography{iclr2025_conference}

\begin{thebibliography}{38}
\providecommand{\natexlab}[1]{#1}
\providecommand{\url}[1]{\texttt{#1}}
\expandafter\ifx\csname urlstyle\endcsname\relax
  \providecommand{\doi}[1]{doi: #1}\else
  \providecommand{\doi}{doi: \begingroup \urlstyle{rm}\Url}\fi

\bibitem[Bao et~al.(2022)Bao, Li, Sun, Zhu, and Zhang]{bao2022estimating}
Fan Bao, Chongxuan Li, Jiacheng Sun, Jun Zhu, and Bo~Zhang.
\newblock Estimating the optimal covariance with imperfect mean in diffusion
  probabilistic models.
\newblock In \emph{ICML}, 2022.

\bibitem[Botchev \& Verwer(2009)Botchev and Verwer]{botchev2009numerical}
Mike~A Botchev and Jan~G Verwer.
\newblock Numerical integration of damped maxwell equations.
\newblock \emph{SIAM Journal on Scientific Computing}, 31\penalty0
  (2):\penalty0 1322--1346, 2009.

\bibitem[Choi et~al.(2020)Choi, Uh, Yoo, and Ha]{choi2020stargan}
Yunjey Choi, Youngjung Uh, Jaejun Yoo, and Jung-Woo Ha.
\newblock Stargan v2: Diverse image synthesis for multiple domains.
\newblock In \emph{CVPR}, 2020.

\bibitem[Deng et~al.(2009)Deng, Dong, Socher, Li, Li, and
  Fei-Fei]{deng2009imagenet}
Jia Deng, Wei Dong, Richard Socher, Li-Jia Li, Kai Li, and Li~Fei-Fei.
\newblock Imagenet: A large-scale hierarchical image database.
\newblock In \emph{CVPR}, 2009.

\bibitem[Dhariwal \& Nichol(2021)Dhariwal and Nichol]{dhariwal2021diffusion}
Prafulla Dhariwal and Alexander Nichol.
\newblock Diffusion models beat {GAN}s on image synthesis.
\newblock In \emph{NeurIPS}, 2021.

\bibitem[Dockhorn et~al.(2022)Dockhorn, Vahdat, and Kreis]{dockhorn2022genie}
Tim Dockhorn, Arash Vahdat, and Karsten Kreis.
\newblock {GENIE}: Higher-order denoising diffusion solvers.
\newblock In \emph{NeurIPS}, 2022.

\bibitem[Heusel et~al.(2017)Heusel, Ramsauer, Unterthiner, Nessler, and
  Hochreiter]{heusel2017gans}
Martin Heusel, Hubert Ramsauer, Thomas Unterthiner, Bernhard Nessler, and Sepp
  Hochreiter.
\newblock Gans trained by a two time-scale update rule converge to a local nash
  equilibrium.
\newblock In \emph{NeurIPS}, 2017.

\bibitem[Ho et~al.(2020)Ho, Jain, and Abbeel]{ho2020denoising}
Jonathan Ho, Ajay Jain, and Pieter Abbeel.
\newblock Denoising diffusion probabilistic models.
\newblock In \emph{NeurIPS}, 2020.

\bibitem[Ho et~al.(2022)Ho, Salimans, Gritsenko, Chan, Norouzi, and
  Fleet]{ho2022video}
Jonathan Ho, Tim Salimans, Alexey Gritsenko, William Chan, Mohammad Norouzi,
  and David~J. Fleet.
\newblock Video diffusion models.
\newblock In \emph{NeurIPS}, 2022.

\bibitem[Kang et~al.(2024)Kang, Choi, Choi, and Han]{kang2024ogdm}
Junoh Kang, Jinyoung Choi, Sungik Choi, and Bohyung Han.
\newblock Observation-guided diffusion probabilistic models.
\newblock In \emph{CVPR}, 2024.

\bibitem[Karras et~al.(2019)Karras, Laine, and Aila]{karras2019style}
Tero Karras, Samuli Laine, and Timo Aila.
\newblock A style-based generator architecture for generative adversarial
  networks.
\newblock In \emph{CVPR}, 2019.

\bibitem[Karras et~al.(2022)Karras, Aittala, Aila, and
  Laine]{karras2022elucidating}
Tero Karras, Miika Aittala, Timo Aila, and Samuli Laine.
\newblock Elucidating the design space of diffusion-based generative models.
\newblock In \emph{NeurIPS}, 2022.

\bibitem[Kim et~al.(2024)Kim, Lai, Liao, Murata, Takida, Uesaka, He, Mitsufuji,
  and Ermon]{kim2024ctm}
Dongjun Kim, Chieh-Hsin Lai, Wei-Hsiang Liao, Naoki Murata, Yuhta Takida,
  Toshimitsu Uesaka, Yutong He, Yuki Mitsufuji, and Stefano Ermon.
\newblock Consistency trajectory models: Learning probability flow ode
  trajectory of diffusion.
\newblock In \emph{ICLR}, 2024.

\bibitem[Krizhevsky \& Hinton(2009)Krizhevsky and
  Hinton]{krizhevsky2009learning}
Alex Krizhevsky and Geoffrey Hinton.
\newblock Learning multiple layers of features from tiny images.
\newblock Technical report, University of Toronto, Toronto, Ontario, 2009.

\bibitem[Lin et~al.(2014)Lin, Maire, Belongie, Bourdev, Girshick, Hays, Perona,
  Ramanan, Doll{'{a} }r, and Zitnick]{cocodataset}
Tsung{-}Yi Lin, Michael Maire, Serge~J. Belongie, Lubomir~D. Bourdev, Ross~B.
  Girshick, James Hays, Pietro Perona, Deva Ramanan, Piotr Doll{'{a} }r, and
  C.~Lawrence Zitnick.
\newblock Microsoft {COCO:} common objects in context.
\newblock \emph{CoRR}, abs/1405.0312, 2014.
\newblock URL \url{http://arxiv.org/abs/1405.0312}.

\bibitem[Liu et~al.(2022)Liu, Ren, Lin, and Zhao]{liu2022pseudo}
Luping Liu, Yi~Ren, Zhijie Lin, and Zhou Zhao.
\newblock Pseudo numerical methods for diffusion models on manifolds.
\newblock In \emph{ICLR}, 2022.

\bibitem[Liu et~al.(2015)Liu, Luo, Wang, and Tang]{liu2015faceattributes}
Ziwei Liu, Ping Luo, Xiaogang Wang, and Xiaoou Tang.
\newblock Deep learning face attributes in the wild.
\newblock In \emph{ICCV}, 2015.

\bibitem[Lu et~al.(2022)Lu, Zhou, Bao, Chen, Li, and Zhu]{lu2022fastode}
Cheng Lu, Yuhao Zhou, Fan Bao, Jianfei Chen, Chongxuan Li, and Jun Zhu.
\newblock {DPM-Solver}: A fast {ODE} solver for diffusion probabilistic model
  sampling in around 10 steps.
\newblock In \emph{NeurIPS}, 2022.

\bibitem[Lu et~al.(2023)Lu, Zhou, Bao, Chen, Li, and Zhu]{lu2023dpm}
Cheng Lu, Yuhao Zhou, Fan Bao, Jianfei Chen, Chongxuan Li, and Jun Zhu.
\newblock Dpm-solver++: Fast solver for guided sampling of diffusion
  probabilistic models.
\newblock In \emph{ICLR}, 2023.

\bibitem[Richards(1997)]{richards1997completed}
Shane~A Richards.
\newblock Completed richardson extrapolation in space and time.
\newblock \emph{Communications in numerical methods in engineering},
  13\penalty0 (7):\penalty0 573--582, 1997.

\bibitem[Richardson(1911)]{richardson1911extra}
Lewis~Fry Richardson.
\newblock Ix. the approximate arithmetical solution by finite differences of
  physical problems involving differential equations, with an application to
  the stresses in a masonry dam.
\newblock \emph{Philosophical Transactions of the Royal Society of London},
  210\penalty0 (459-470):\penalty0 357--357, 1911.
\newblock \doi{https://doi.org/10.1098/rsta.1911.0009}.

\bibitem[Rombach et~al.(2022)Rombach, Blattmann, Lorenz, Esser, and
  Ommer]{rombach2022high}
Robin Rombach, Andreas Blattmann, Dominik Lorenz, Patrick Esser, and Bj{\"o}rn
  Ommer.
\newblock High-resolution image synthesis with latent diffusion models.
\newblock In \emph{CVPR}, 2022.

\bibitem[Salimans \& Ho(2022)Salimans and Ho]{salimans2022progressive}
Tim Salimans and Jonathan Ho.
\newblock Progressive distillation for fast sampling of diffusion models.
\newblock In \emph{ICLR}, 2022.

\bibitem[Singer et~al.(2022)Singer, Polyak, Hayes, Yin, An, Zhang, Hu, Yang,
  Ashual, Gafni, Parikh, Gupta, and Taigman]{singer2022make}
Uriel Singer, Adam Polyak, Thomas Hayes, Xi~Yin, Jie An, Songyang Zhang, Qiyuan
  Hu, Harry Yang, Oron Ashual, Oran Gafni, Devi Parikh, Sonal Gupta, and Yaniv
  Taigman.
\newblock {Make-A-Video}: Text-to-video generation without text-video data.
\newblock In \emph{ICLR}, 2022.

\bibitem[Song et~al.(2021{\natexlab{a}})Song, Meng, and
  Ermon]{song2021denoising}
Jiaming Song, Chenlin Meng, and Stefano Ermon.
\newblock Denoising diffusion implicit models.
\newblock In \emph{ICLR}, 2021{\natexlab{a}}.

\bibitem[Song et~al.(2021{\natexlab{b}})Song, Sohl-Dickstein, Kingma, Kumar,
  Ermon, and Poole]{song2021scorebased}
Yang Song, Jascha Sohl-Dickstein, Diederik~P Kingma, Abhishek Kumar, Stefano
  Ermon, and Ben Poole.
\newblock Score-based generative modeling through stochastic differential
  equations.
\newblock In \emph{ICLR}, 2021{\natexlab{b}}.

\bibitem[Song et~al.(2023)Song, Dhariwal, Chen, and
  Sutskever]{song2023consistency}
Yang Song, Prafulla Dhariwal, Mark Chen, and Ilya Sutskever.
\newblock Consistency models.
\newblock In \emph{ICML}, 2023.

\bibitem[S{\"u}li \& Mayers(2003)S{\"u}li and Mayers]{suli2003introduction}
Endre S{\"u}li and David Mayers.
\newblock \emph{An Introduction to Numerical Analysis}.
\newblock Cambridge University Press, 1 edition, 2003.
\newblock ISBN 0-521-00794-1.

\bibitem[Timothy(2017)]{timothy2017numerical}
Sauer Timothy.
\newblock \emph{Numerical Analysis}.
\newblock Pearson, 3 edition, 2017.
\newblock ISBN 2017028491, 9780134696454, 013469645X.

\bibitem[Wang et~al.(2023)Wang, Yuan, Chen, Zhang, Wang, and
  Zhang]{wang2023modelscope}
Jiuniu Wang, Hangjie Yuan, Dayou Chen, Yingya Zhang, Xiang Wang, and Shiwei
  Zhang.
\newblock {ModelScope} text-to-video technical report.
\newblock \emph{arXiv preprint arXiv:2308.06571}, 2023.

\bibitem[Xiao et~al.(2022)Xiao, Kreis, and Vahdat]{xiao2022DDGAN}
Zhisheng Xiao, Karsten Kreis, and Arash Vahdat.
\newblock Tackling the generative learning trilemma with denoising diffusion
  {GAN}s.
\newblock In \emph{ICLR}, 2022.

\bibitem[Yu et~al.(2015)Yu, Seff, Zhang, Song, Funkhouser, and
  Xiao]{yu2015lsun}
Fisher Yu, Ari Seff, Yinda Zhang, Shuran Song, Thomas Funkhouser, and Jianxiong
  Xiao.
\newblock {LSUN}: Construction of a large-scale image dataset using deep
  learning with humans in the loop.
\newblock \emph{arXiv preprint arXiv:1506.03365}, 2015.

\bibitem[Zeng et~al.(2022)Zeng, Vahdat, Williams, Gojcic, Litany, Fidler, and
  Kreis]{zeng2022lion}
Xiaohui Zeng, Arash Vahdat, Francis Williams, Zan Gojcic, Or~Litany, Sanja
  Fidler, and Karsten Kreis.
\newblock {LION}: Latent point diffusion models for {3D} shape generation.
\newblock In \emph{NeurIPS}, 2022.

\bibitem[Zhang et~al.(2023)Zhang, Kenta, and Kleijn]{zhang2023lookahead}
Guoqiang Zhang, Niwa Kenta, and W.~Bastiaan Kleijn.
\newblock Lookahead diffusion probabilistic models for refining mean
  estimation.
\newblock In \emph{CVPR}, 2023.

\bibitem[Zhang et~al.(2024)Zhang, Kenta, and Kleijn]{zhang2024iia}
Guoqiang Zhang, Niwa Kenta, and W.~Bastiaan Kleijn.
\newblock On accelerating diffusion-based sampling process via improved
  integration approximation.
\newblock In \emph{ICLR}, 2024.

\bibitem[Zhang \& Chen(2023)Zhang and Chen]{zhang2023fast}
Qinsheng Zhang and Yongxin Chen.
\newblock Fast sampling of diffusion models with exponential integrator.
\newblock In \emph{ICLR}, 2023.

\bibitem[Zhou et~al.(2022)Zhou, Wang, Yan, Lv, Zhu, and Feng]{zhou2022magic}
Daquan Zhou, Weimin Wang, Hanshu Yan, Weiwei Lv, Yizhe Zhu, and Jiashi Feng.
\newblock {MagicVideo}: Efficient video generation with latent diffusion
  models.
\newblock \emph{arXiv preprint arXiv:2211.11018}, 2022.

\bibitem[Zlatev et~al.(2010)Zlatev, Farag{\'o}, and
  Havasi]{zlatev2010stability}
Zahari Zlatev, Istv{\'a}n Farag{\'o}, and {\'A}gnes Havasi.
\newblock Stability of the richardson extrapolation applied together with the
  $\theta$-method.
\newblock \emph{Journal of Computational and Applied Mathematics}, 235\penalty0
  (2):\penalty0 507--517, 2010.

\end{thebibliography}
\bibliographystyle{iclr2025_conference}

\appendix

\clearpage
\section{Justification on \cref{eq:ode_e2}}
\label{app:justification}

Assuming a uniform grid as in the context of conventional Richardson extrapolation in \cref{subsec:richardson_extrapolation}, and the ODE solver with $O(h^{p+1})$ of local truncation error formula, we have
\begin{align}
    V^* &= V(h) + ch^{p} + O(h^{p+1}) \quad \text{and} \label{eq:local_richard_1}\\
    V^* &= V(\frac{h}{k}) + c(\frac{h}{k})^{p} + O(h^{p+1}) \label{eq:local_richard_k}    
\end{align}
since we expect the $O(h^p)$ of the global truncation error.
Then, we have the following extrapolated solution with a truncation error of $O(h^{p+1})$ by Richardson extrapolation (\cref{eq:ric_extra_sol}):
\begin{align}
    \tilde{V}(h,k) = \cfrac{k^p V(h/k) - V(h)}{k^p -1}.
    \label{eq:ric_extra_sol_app}
\end{align}
Now, considering the case of \cref{eq:ode_e1,eq:ode_e2} with uniform discretization, we have
\begin{align}
    V^* &= V(h) + c’h^{p+1} + O(h^{p+2}) \quad \text{and} \label{eq:local_rx_1}\\
    V^* &= V(\frac{h}{k}) + kc’(\frac{h}{k})^{p+1} + O(h^{p+2}), \label{eq:local_rx_k}
\end{align}
each correspondingly. Then, the extrapolated solution $\tilde{V}_{\text{ours}}$ obtained from solving linear system of \cref{eq:local_rx_1,eq:local_rx_k} becomes
\begin{align}
    \tilde{V}_{\text{ours}} = \cfrac{k^p V(h/k) - V(h)}{k^p -1},
    \label{eq:our_extra_sol_app}
\end{align}
which turns out to be exactly the same as \cref{eq:ric_extra_sol_app}.
Thus, we believe our approach can be considered to employ assumptions shared by those used in common practices of Richardson extrapolation and also can reduce global errors which is backed by experimental results as well.

\section{More quantitative results}
\label{app:quantitative}
\subsection{Comparison with DEIS}
\label{app:deis}
We compare DEIS variants~\citep{zhang2023fast}, DPM-Solvers, and RX-DPMs applied to DPM-Solvers on class-conditioned ImageNet (64$\times$64) in \cref{tab:deis}.
The results clearly demonstrate that our method outperforms all other approaches across all NFEs.

\subsection{DPMs with optimal covariances}
\label{app:sde_ddim}
Although our method is designed for ODE solvers, we also conduct experiments with SN-DPM and NPR-DPM~\citep{bao2022estimating} on CIFAR-10~\citep{krizhevsky2009learning} and CelebA datasets~\citep{liu2015faceattributes} to evaluate its performance when applied to SDE solvers.
We use the official codes and provided pretrained models as described in \cref{subsec:implement}.
SN-DPM and NPR-DPM are two different models that correct the imperfect mean prediction in the reverse process of existing models through optimal covariance learning. 

To incorporate our method into stochastic sampling, we decompose it into a deterministic sampling component and a stochastic component, and apply our method to the deterministic sampling part.
Specifically, within each $k$-step interval, we execute RX-DPM algorithm using the deterministic sampling component and then add the stochasticity term afterward. 
In this way, our method uses the stochasticity of the baseline sampling method in a limited manner compared to the baseline sampling with the same NFEs.
For a better understanding of our implementation, we provide the diagrams of the proposed method in \cref{app:diagram}.

In \cref{tab:rx_ea}, we show the results on NPR-DPM and SN-DPM along with the vanilla DDIM and LA-DPM~\citep{zhang2023lookahead}, which is another extrapolation-based sampling method. 
We observe that our method outperforms the compared methods in most cases, although a performance degradation is noted with RX-SN-DDIM on the CIFAR-10 dataset. 
This implies that our approach of solving the ODE might offset the benefits of the model's optimization. 
Despite this, we observe significant performance improvements in the most extreme case, $\text{NFEs}=10$.
\begin{table}[t]
    \centering
    \setlength{\tabcolsep}{3mm}
    \caption{Comparisons of DEIS variants, DPM-Solvers and RX-DPMs applied to DPM-Solvers on class-conditioned ImageNet (64$\times$64). All results of DEIS and DPM-Solvers are copied from DEIS~\citep{zhang2023fast} except for the result of DPM-Solver-3 with $\text{NFEs}=9$.  }
    \vspace{-1mm}
    \scalebox{0.8}{
    \begin{tabular}{lccccc}
        \toprule
        & \multicolumn{5}{c}{NFEs} \\
        \cmidrule(lr){2-6}
        Method & 9 & 10 & 12 & 18 & 30 \\
        \midrule
        tAB-DEIS & --& 6.65 & 3.99 & 3.21 & 2.81 \\
        $\rho$AB-DEIS &-- & 9.28 & 6.46 & 3.74 & 2.87 \\
        \midrule
        DPM-Solver-2 & --& 7.93 & 5.36 & 3.63 & 3.00 \\
        $\rho$Mid-DEIS & --& 9.12 & 6.78 & 4.00 & 2.99 \\
        RX-DPM-Solver-2 & --& \B6.11 & 5.61 & 3.64 & 2.93 \\
        \midrule
        DPM-Solver-3 & 7.45 & -- &5.02 & 3.18 & 2.84 \\
        $\rho$Kutta-DEIS & -- & -- &13.12 & 3.63 & 2.82 \\
        RX-DPM-Solver-3 & \B7.08 && \B3.90 & \B2.36 & \B2.18 \\
        \bottomrule
    \end{tabular}
    }
    \label{tab:deis}
\end{table}
\begin{table}[t]
    \centering
    \caption{FID scores for CIFAR-10 and CelebA on DDIM, NPR-DDIM and SN-DDIM models. 
    The values for each baseline and LA-DDIM results are copied from \cite{zhang2023lookahead}.}
    \vspace{-1mm}
    \label{tab:rx_ea}
    \renewcommand{\arraystretch}{1.0}
    \setlength{\tabcolsep}{3mm}
    \scalebox{0.8}{
        \begin{tabular}{lcccccccccc}
            \toprule
            Dataset & & & \multicolumn{3}{c}{CIFAR-10 (32$\times$32)} & & \multicolumn{3}{c}{CelebA (64$\times$64)} \\
            \cmidrule{1-2} \cmidrule{4-6} \cmidrule{8-10}

            Method ~~\textbackslash~~ NFEs & & & 10 & 25 & 50 &  & 10 & 25 & 50 \\
            \cmidrule{1-2} \cmidrule{4-6} \cmidrule{8-10}

            DDIM &  &  & 21.31 & 10.70 & 7.74 &  & 20.54 & 13.45 & 9.33 \\
            RX-DDIM &  &  & \B14.78 & \B 8.42 & \B6.30 &  & \B18.31 & \B10.54 & \B6.88 \\
            \hline
            NPR-DDIM &  &  & 13.40 & 5.43 & 3.99 &  & 14.94 & 9.18 & 6.17 \\
            LA-NPR-DDIM &  &  & 10.74 & 4.71 & 3.64 &  & 14.25 & 8.83 & 5.67 \\
            RX-NPR-DDIM &  &  & \ \ \B6.35 & \B3.92 & \B3.34 &  & \B11.58 & \B6.61 & \B3.98 \\
            \hline
            SN-DDIM &  &  & 12.19 & 4.28 & 3.39 &  & 10.17 & 5.62 & 3.90 \\
            LA-SN-DDIM &  &  & \ \ 8.48 & \B3.15 & \B2.93 &  &\ \  8.05 & 4.56 & 2.93  \\
            RX-SN-DDIM &  &  & \ \ \B7.50 & 5.12 & 4.40 &  & \ \ \B5.20 & \B2.72 & \B2.25 \\
            \bottomrule
        \end{tabular}}
        \vspace{-2mm}
\end{table}

\section{Diagrams}
\label{app:diagram}
\begin{figure}[t]
    \centering
    \scalebox{1.0}{
    \begin{tabular}{c}
    \includegraphics[width=0.4\textwidth]{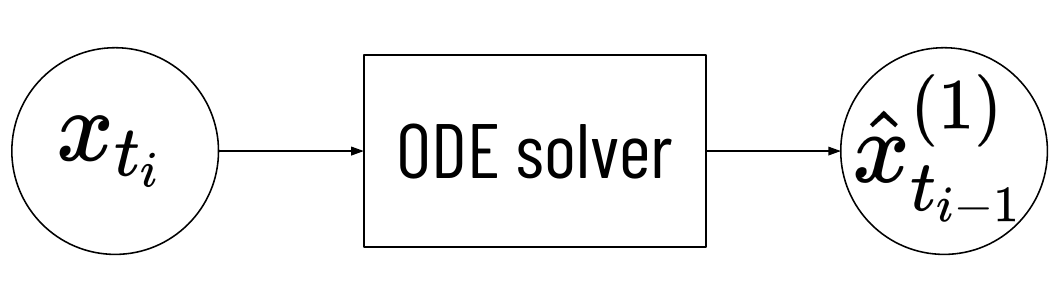} \\
    (a) One step of an ODE solver
    \vspace{2mm}
    \\
    \includegraphics[width=0.7\textwidth]{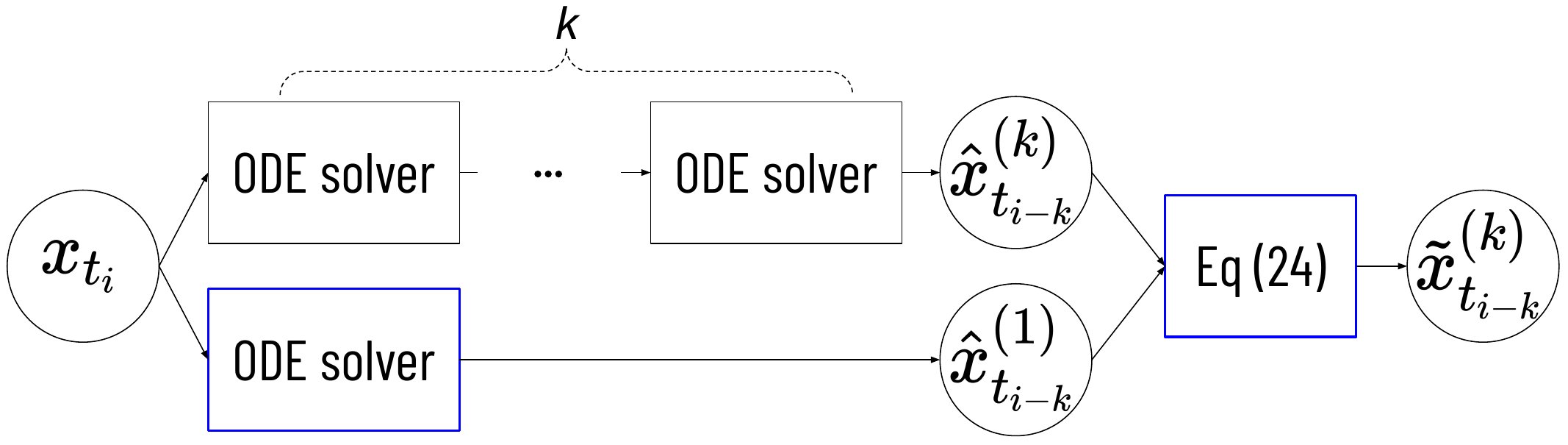} \\
    (b) $k$ steps of RX-DPM applied to an ODE solver
    \vspace{2mm}
    \\
    \includegraphics[width=0.85\textwidth]{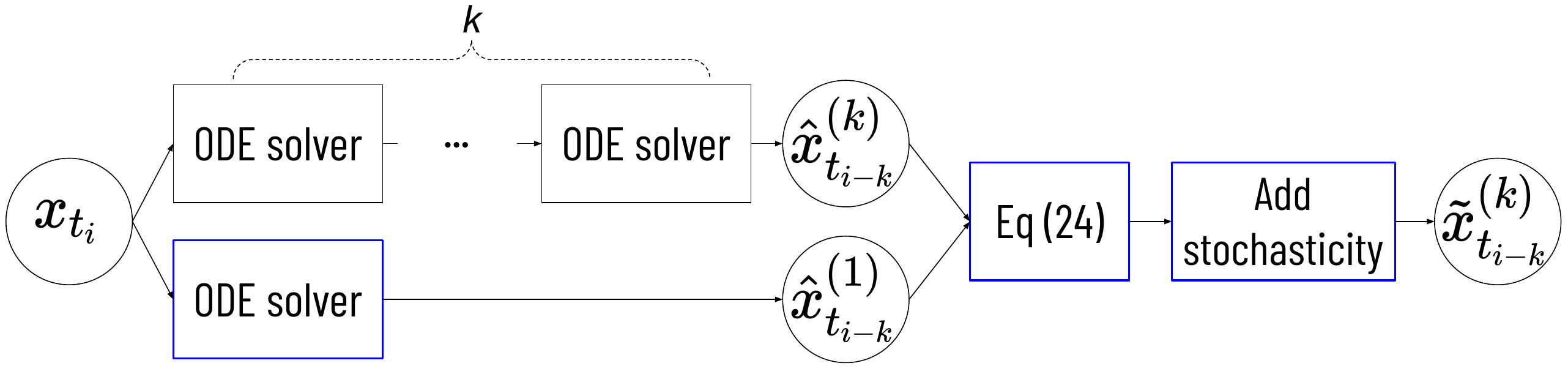} \\
    (c) $k$ steps of RX-DPM applied to an SDE solver
    \end{tabular}
}
\caption{Digarams of the baseline and the proposed sampling methods. 
The blue-bordered boxes in (b) and (c) indicate that the corresponding operation does not require network evaluation. The ODE solver in (c) refers to the deterministic sampling component of the SDE solver.}
\label{fig:diagram}
\end{figure}

\cref{fig:diagram} compares the diagrams of an ODE solver, the proposed method with an ODE solver, and the proposed method with an SDE solver. 

\section{Computational cost}
\label{app:time}

We compare the computational time and GPU memory usage of the Euler method and RX-Euler using the EDM backbone in \cref{tab:time,tab:gpu}, respectively.
For measurements, we set the batch size to 128 and use 10-step sampling on an A6000 GPU. 
The average runtime per batch is measured for computational time.
The additional operations introduced by our method, which consist of linear combinations of precomputed values, result in negligible computational overhead compared to the time required for the network forward pass. 
Furthermore, as the model size increases, the relative overhead diminishes (\eg, only 0.11\% increase for ImageNet class-conditional sampling).
Similarly, GPU memory usage increases slightly with RX-Euler, primarily due to the storage of previous predictions.
However, this increase is minimal and decreases as model or data size increases, showing a similar trend to that observed with computational time.
\section{Limitations}
\label{sec:limitation}
As our method is primarily designed for an ODE solver, to integrate it with an SDE solver (or a stochastic sampling method), we partially apply the stochasticity component of the SDE solver as demonstrated in \cref{app:sde_ddim}. 
Consequently, in some cases, the effectiveness is offset because the full effects of stochasticity are not captured. 
However, in scenarios where NFE is very limited, which are of greater interest to us, the combined effect of RX-DPM and stochastic sampling has been empirically shown to be highly beneficial. 
We leave the development of methods that can perform better in more general cases for future work. 
Additionally, in the extension of the RX-Euler algorithm to a higher-order solver in Section \cref{subsec:rx-ode}, there remains room for improvement since we impose assumptions about linear error propagation. 
We believe that relaxing these assumptions or deriving more accurate equations could further enhance the performance.

\begin{table}[t]
    \centering
    \setlength{\tabcolsep}{4mm}
    \caption{Comparison of per-batch computation times between the Euler method and RX-Euler with the EDM backbone. The reported values represent the average runtime across 100 measurements (in seconds).}
    \scalebox{0.8}{
    \begin{tabular}{lccc}
        \toprule
        & CIFAR-10 cond. (32x32) & FFHQ (64x64)      & ImageNet cond. (64x64) \\
        \midrule
        Euler    & 1.737 $\pm$ 0.028      & 3.895 $\pm$ 0.023 & 6.436 $\pm$ 0.033      \\
        RX-Euler & 1.743 $\pm$ 0.031      & 3.903 $\pm$ 0.025 & 6.443 $\pm$ 0.039     \\
        \bottomrule
    \end{tabular}
    }
    \label{tab:time}
\end{table}
\begin{table}[H]
    \centering
    \setlength{\tabcolsep}{4mm}
    \caption{Comparison of GPU memory usage (MiB) during inference between the Euler method and RX-Euler with the EDM backbone.}
    \scalebox{0.8}{
    \begin{tabular}{lccc}
        \toprule
        & CIFAR-10 cond. (32x32) & FFHQ (64x64)      & ImageNet cond. (64x64) \\
        \midrule
        Euler    & 2929	& 9433 & 12661 \\
        RX-Euler & 3033 & 9477 & 12705 \\
        \bottomrule
    \end{tabular}
    }
    \label{tab:gpu}
\end{table}

\section{Qualitative results}
\label{app:qual}
\begin{figure}
    \centering
    \renewcommand{\arraystretch}{0.7}
    \scalebox{0.98}{
    \setlength{\tabcolsep}{1pt} \hspace{-0.5mm}
        \hspace{-3mm}
    \begin{tabular}{cc}
        \includegraphics[width=0.48\textwidth]{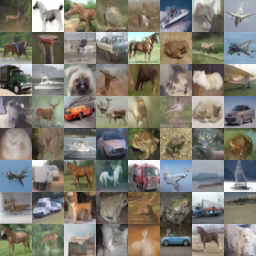} &
        \includegraphics[width=0.48\textwidth]{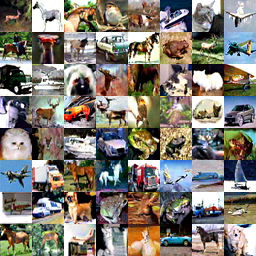}\\
        Euler ($\text{NFEs}=10$, FID: 15.88) &
        EDM ($\text{NFEs}=11$, FID: 14.46) \\
        &\\        
        \includegraphics[width=0.48\textwidth]{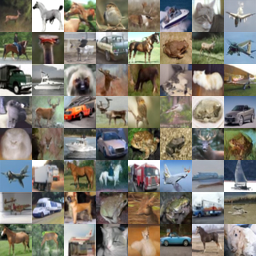} &
        \includegraphics[width=0.48\textwidth]{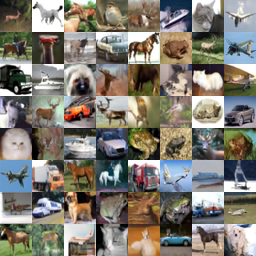}\\
        RX-Euler ($\text{NFEs}=10$, FID 4.35) &
        RX+EDM ($\text{NFEs}=10$, FID 4.26)
    \end{tabular}
    }
    \caption{Qualitative results of CIFAR-10 of different sampling methods with EDM backbone.}
    \label{fig:qual_cifar}
\end{figure}

\begin{figure}
    \centering
    \renewcommand{\arraystretch}{0.7}
    \scalebox{0.98}{
    \setlength{\tabcolsep}{1pt} \hspace{-0.5mm}
        \hspace{-3mm}
    \begin{tabular}{cc}

        \includegraphics[width=0.48\textwidth]{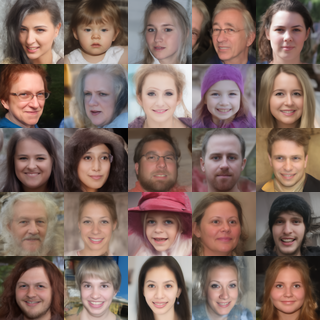} &
        \includegraphics[width=0.48\textwidth]{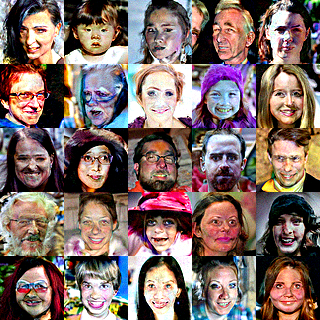}\\
        Euler ($\text{NFEs}=9$, FID: 24.27) &
        EDM ($\text{NFEs}=9$, FID: 57.12) \\        
        & \\
        \includegraphics[width=0.48\textwidth]{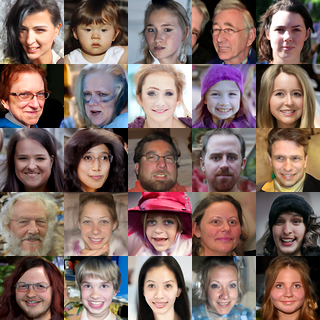} &
        \includegraphics[width=0.48\textwidth]{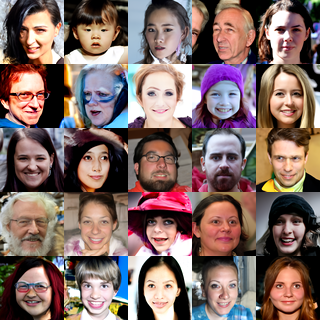}\\
        RX-Euler ($\text{NFEs}=8$, FID: 12.94) &
        RX+EDM ($\text{NFEs}=9$, FID: 6.83)
    \end{tabular}
    }
    \caption{Qualitative results of FFHQ of different sampling methods with EDM backbone.}
    \label{fig:qual_ffhq1}
\end{figure}

\begin{figure}
    \centering
    \renewcommand{\arraystretch}{0.7}
    \scalebox{0.98}{
    \setlength{\tabcolsep}{1pt} \hspace{-0.5mm}
        \hspace{-3mm}
    \begin{tabular}{cc}

        \includegraphics[width=0.48\textwidth]{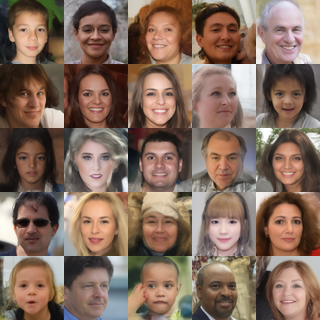} &
        \includegraphics[width=0.48\textwidth]{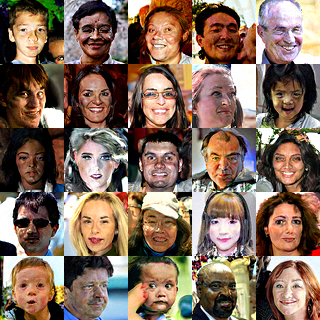}\\
        Euler ($\text{NFEs}=11$, FID: 18.30) &
        EDM ($\text{NFEs}=11$, FID: 29.34) \\        
        & \\
        \includegraphics[width=0.48\textwidth]{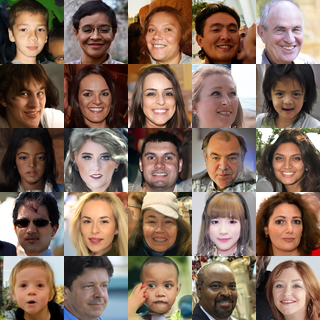} &
        \includegraphics[width=0.48\textwidth]{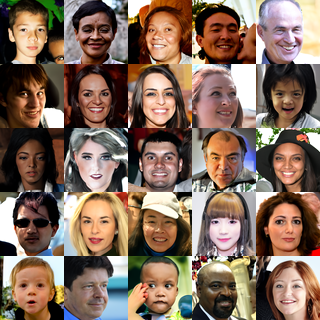}\\
        RX-Euler ($\text{NFEs}=10$, FID: 7.20) &
        RX+EDM ($\text{NFEs}=11$, FID: 5.50)
    \end{tabular}
    }
    \caption{Qualitative results of FFHQ of different sampling methods with EDM backbone.}
    \label{fig:qual_ffhq2}
\end{figure}

\begin{figure}
    \centering
    \renewcommand{\arraystretch}{0.7}
    \scalebox{0.98}{
    \setlength{\tabcolsep}{1pt} \hspace{-0.5mm}
        \hspace{-3mm}
    \begin{tabular}{cc}

        \includegraphics[width=0.48\textwidth]{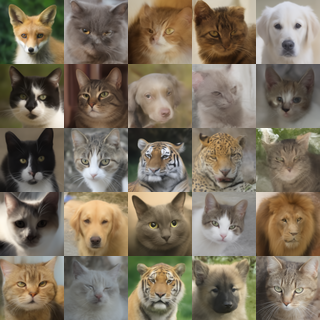} &
        \includegraphics[width=0.48\textwidth]{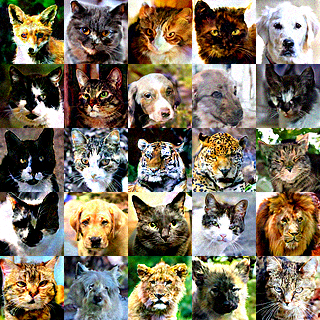}\\
        Euler ($\text{NFEs}=9$, FID: 11.81) &
        EDM ($\text{NFEs}=9$, FID: 27.92) \\
        & \\
        \includegraphics[width=0.48\textwidth]{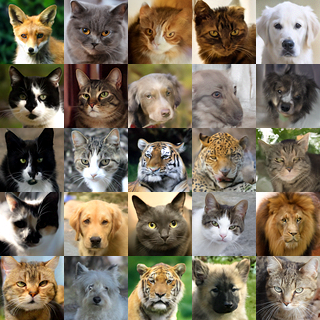} &
        \includegraphics[width=0.48\textwidth]{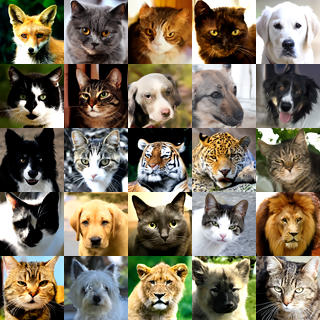}\\
        RX-Euler ($\text{NFEs}=8$, FID: 6.49) &
        RX+EDM ($\text{NFEs}=9$, FID: 3.88)
    \end{tabular}
    }
    \caption{Qualitative results of AFHQv2 of different sampling methods with EDM backbone.}
    \label{fig:qual_afhq1}
\end{figure}

\begin{figure}
    \centering
    \renewcommand{\arraystretch}{0.7}
    \scalebox{0.98}{
    \setlength{\tabcolsep}{1pt} \hspace{-0.5mm}
        \hspace{-3mm}
    \begin{tabular}{cc}

        \includegraphics[width=0.48\textwidth]{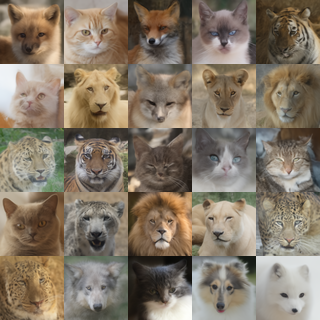} &
        \includegraphics[width=0.48\textwidth]{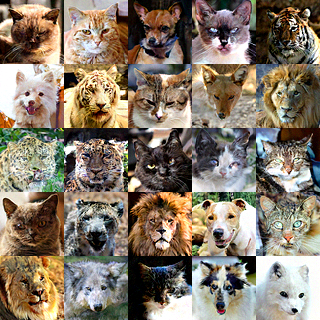}\\
        Euler ($\text{NFEs}=11$, FID: 8.94) &
        EDM ($\text{NFEs}=11$, FID: 13.61) \\
        & \\
        \includegraphics[width=0.48\textwidth]{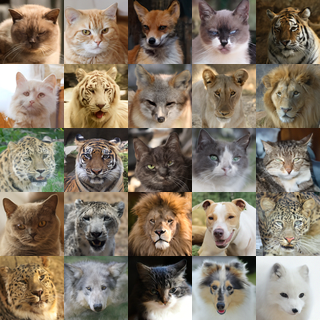} &
        \includegraphics[width=0.48\textwidth]{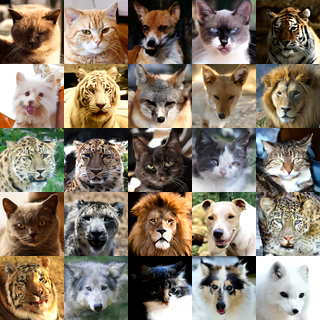}\\
        RX-Euler ($\text{NFEs}=10$, FID: 4.02) &
        RX+EDM ($\text{NFEs}=10$, FID: 3.72)
    \end{tabular}
    }
    \caption{Qualitative results of AFHQv2 of different sampling methods with EDM backbone.}
    \label{fig:qual_afhq2}
\end{figure}

\begin{figure}
    \centering
    \renewcommand{\arraystretch}{0.7}
    \scalebox{0.98}{
    \setlength{\tabcolsep}{1pt} \hspace{-0.5mm}
        \hspace{-3mm}
    \begin{tabular}{cc}

        \includegraphics[width=0.48\textwidth]{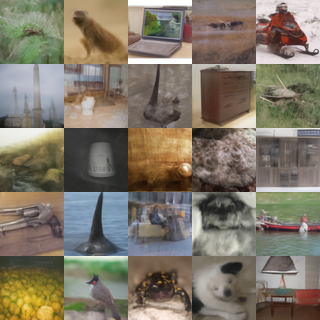} &
        \includegraphics[width=0.48\textwidth]{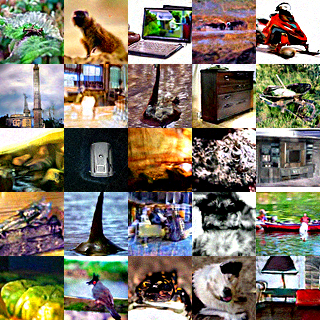}\\
        Euler ($\text{NFEs}=9$, FID: 20.49) &
        EDM ($\text{NFEs}=9$, FID: 35.34) \\
        & \\
        \includegraphics[width=0.48\textwidth]{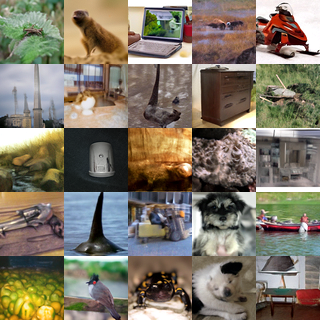} &
        \includegraphics[width=0.48\textwidth]{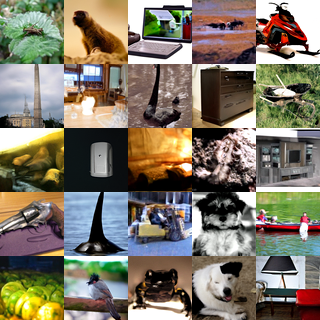}\\
        RX-Euler ($\text{NFEs}=8$, FID: 12.38) &
        RX+EDM ($\text{NFEs}=9$, FID: 7.94)
    \end{tabular}
    }
    \caption{Qualitative results of ImageNet of different sampling methods with EDM backbone.}
    \label{fig:qual_imagenet1}
\end{figure}

\begin{figure}
    \centering
    \renewcommand{\arraystretch}{0.7}
    \scalebox{0.98}{
    \setlength{\tabcolsep}{1pt} \hspace{-0.5mm}
        \hspace{-3mm}
    \begin{tabular}{cc}

        \includegraphics[width=0.48\textwidth]{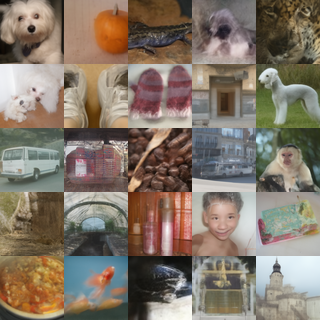} &
        \includegraphics[width=0.48\textwidth]{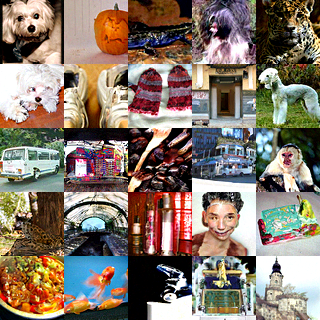}\\
        Euler ($\text{NFEs}=11$, FID: 14.84) &
        EDM ($\text{NFEs}=11$, FID: 15.35) \\
        & \\
        \includegraphics[width=0.48\textwidth]{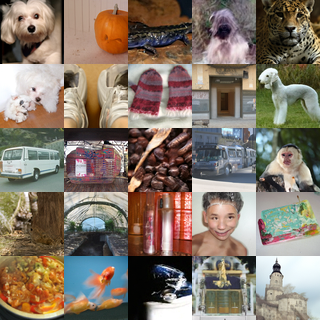} &
        \includegraphics[width=0.48\textwidth]{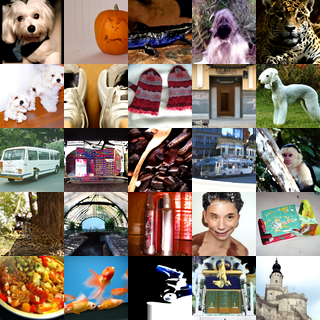}\\
        RX-Euler ($\text{NFEs}=10$, FID: 6.95) &
        RX+EDM ($\text{NFEs}=11$, FID: 5.83)
    \end{tabular}
    }
    \caption{Qualitative results of ImageNet of different sampling methods with EDM backbone.}
    \label{fig:qual_imagenet2}
\end{figure}

\begin{figure}
    \centering
    \renewcommand{\arraystretch}{0.7}
    \scalebox{0.98}{
    \setlength{\tabcolsep}{1pt} \hspace{-0.5mm}
        \hspace{-3mm}
    \begin{tabular}{ccccc}
    & $\text{NFEs}=15$ & $\text{NFEs}=20$ & $\text{NFEs}=30$ & $\text{NFEs}=50$ \\ 
    \rotatebox{90}{\hspace{12mm}{DDIM}}~ &
    \includegraphics[width=0.24\textwidth]{fig/sd/COCO_val2014_000000475529/ddim15.png} &
    \includegraphics[width=0.24\textwidth]{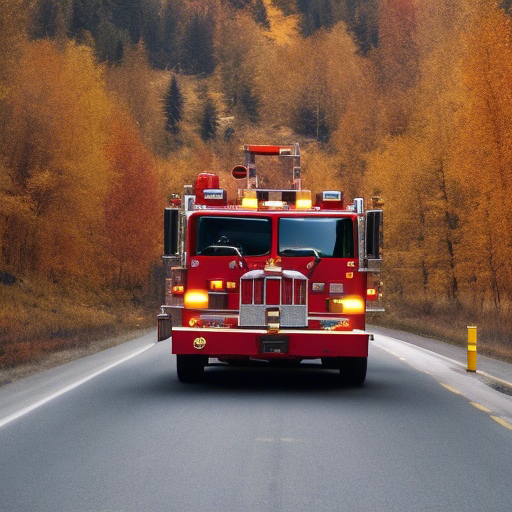} &
    \includegraphics[width=0.24\textwidth]{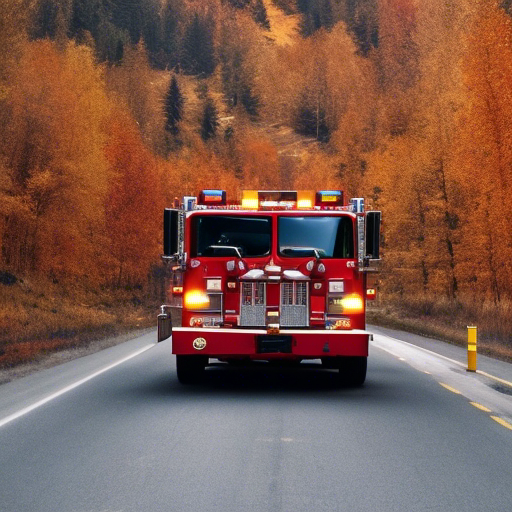} &
    \includegraphics[width=0.24\textwidth]{fig/sd/COCO_val2014_000000475529/ddim50.png} \\
    \rotatebox{90}{\hspace{9mm}{RX-DDIM}}~ & 
    \includegraphics[width=0.24\textwidth]{fig/sd/COCO_val2014_000000475529/ddim15_2.png} &
    \includegraphics[width=0.24\textwidth]{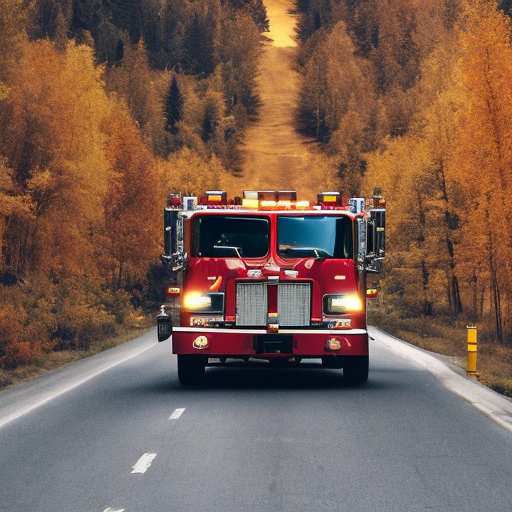} &
    \includegraphics[width=0.24\textwidth]{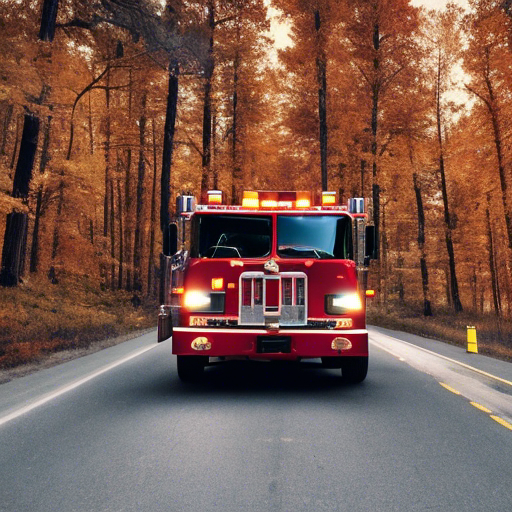} &
    \includegraphics[width=0.24\textwidth]{fig/sd/COCO_val2014_000000475529/ddim50_2.png} \\
    &\multicolumn{4}{c}{\textit{``A red fire truck driving down a road.''}}\vspace{1.5mm} \\
    \rotatebox{90}{\hspace{12mm}{DDIM}}~ &
    \includegraphics[width=0.24\textwidth]{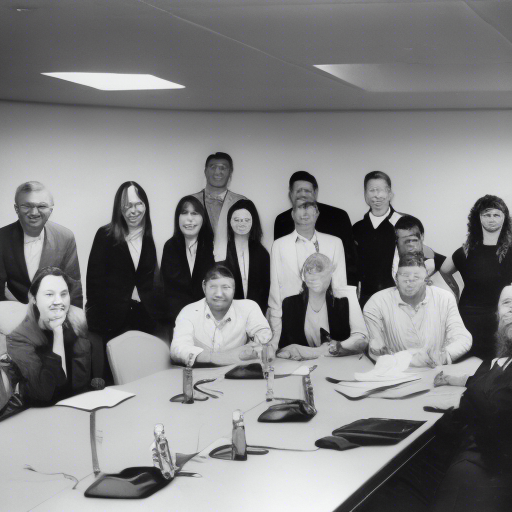} &
    \includegraphics[width=0.24\textwidth]{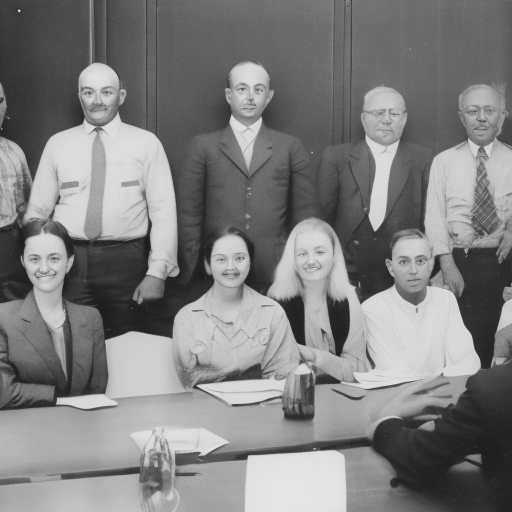} &
    \includegraphics[width=0.24\textwidth]{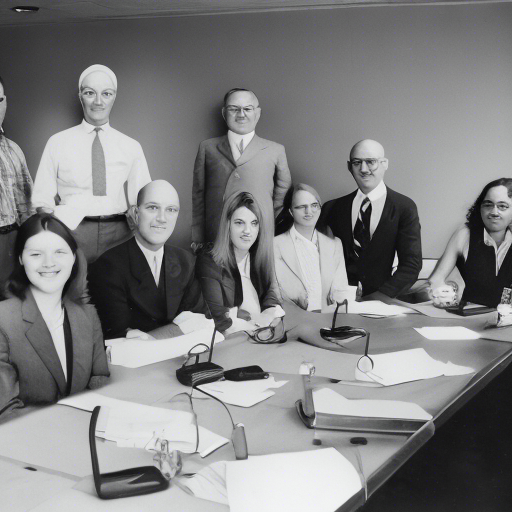} &
    \includegraphics[width=0.24\textwidth]{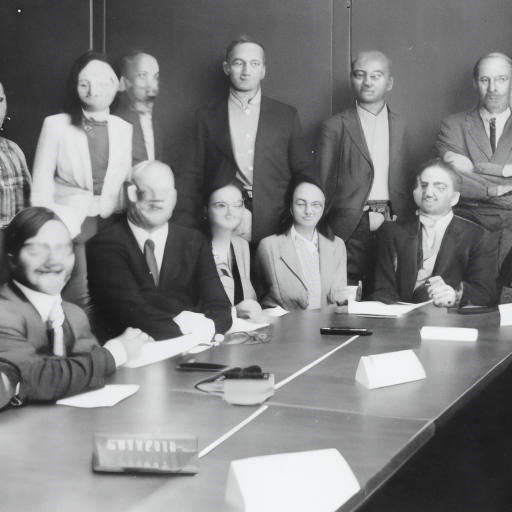} \\
    \rotatebox{90}{\hspace{9mm}{RX-DDIM}}~ & 
    \includegraphics[width=0.24\textwidth]{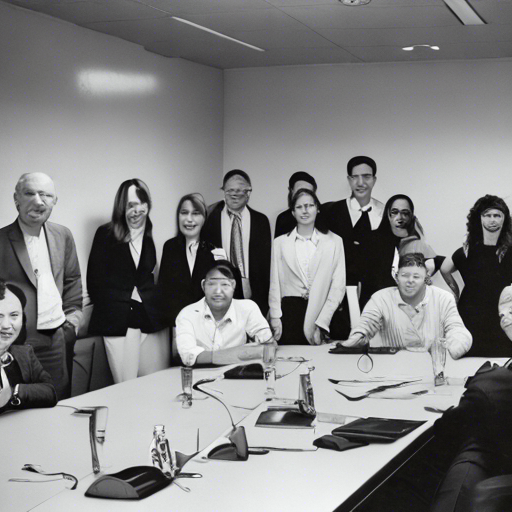} &
    \includegraphics[width=0.24\textwidth]{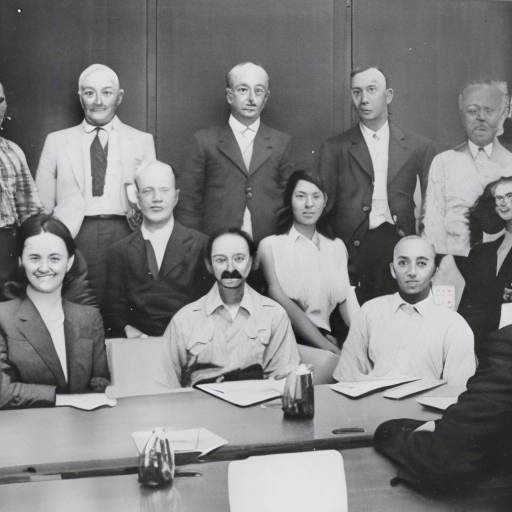} &
    \includegraphics[width=0.24\textwidth]{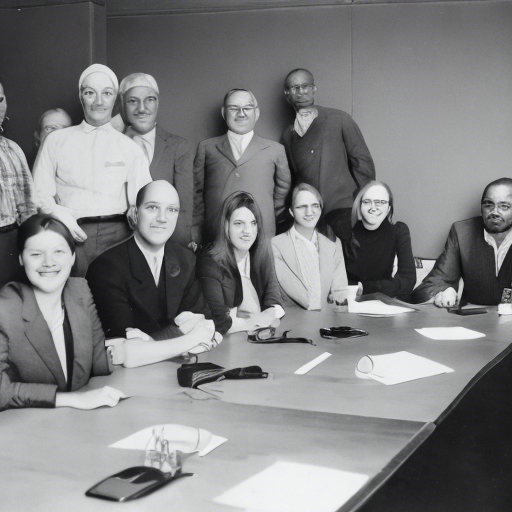} &
    \includegraphics[width=0.24\textwidth]{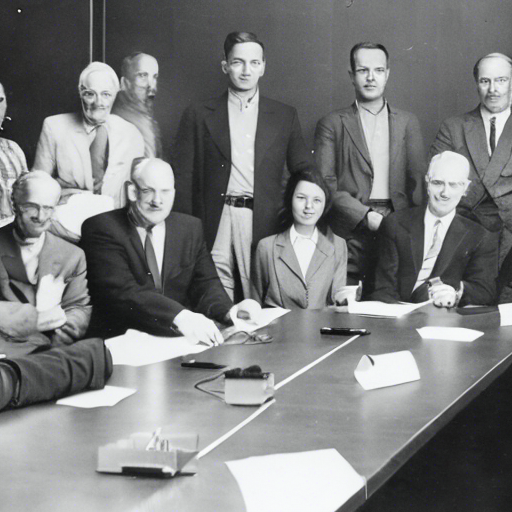} \\
    &\multicolumn{4}{c}{\textit{``Men and women are posing for a photo in a conference room.''}}\vspace{1.5mm} \\
    \rotatebox{90}{\hspace{12mm}{DDIM}}~ &
    \includegraphics[width=0.24\textwidth]{fig/sd/COCO_val2014_000000079565/ddim15.png} &
    \includegraphics[width=0.24\textwidth]{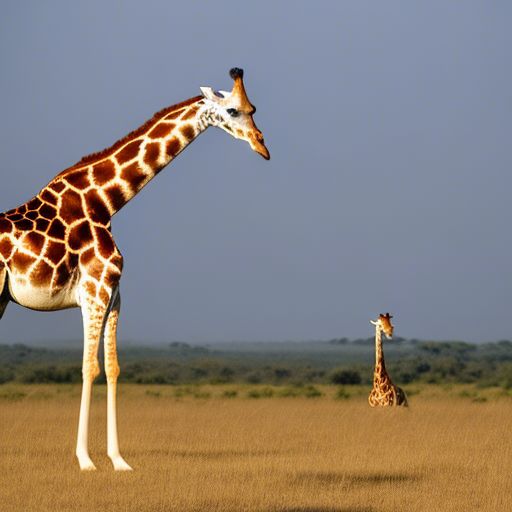} &
    \includegraphics[width=0.24\textwidth]{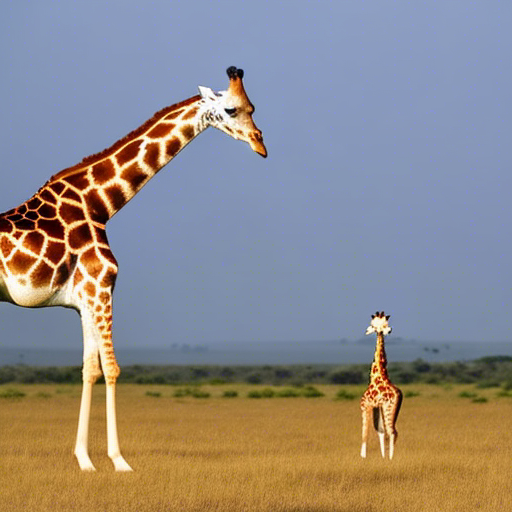} &
    \includegraphics[width=0.24\textwidth]{fig/sd/COCO_val2014_000000079565/ddim50.png} \\
    \rotatebox{90}{\hspace{9mm}{RX-DDIM}}~ & 
    \includegraphics[width=0.24\textwidth]{fig/sd/COCO_val2014_000000079565/ddim15_2.png} &
    \includegraphics[width=0.24\textwidth]{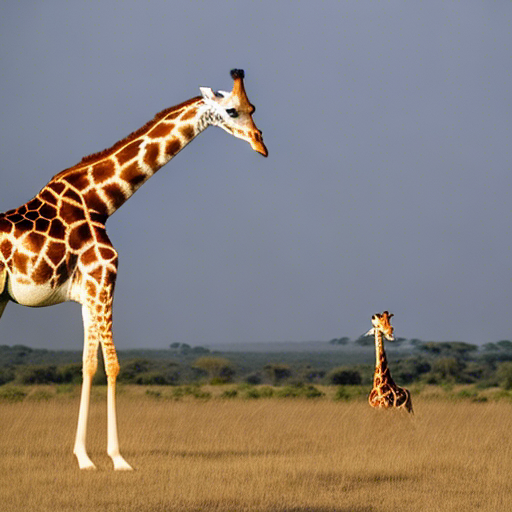} &
    \includegraphics[width=0.24\textwidth]{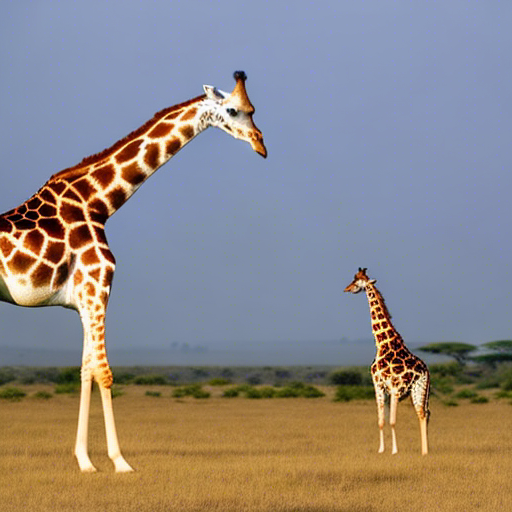} &
    \includegraphics[width=0.24\textwidth]{fig/sd/COCO_val2014_000000079565/ddim50_2.png} \\
    &\multicolumn{4}{c}{\textit{``A pair of giraffe standing in a big open area.''}}\vspace{1.5mm} 
    \end{tabular}
    }
    \caption{Qualitative results on Stable Diffusion V2 of DDIM and RX-DDIM.}
    \label{fig:rx_sd1}
\end{figure}

\begin{figure}
\centering
\renewcommand{\arraystretch}{0.7}
\scalebox{0.98}{
\setlength{\tabcolsep}{1pt} \hspace{-0.5mm}
    \hspace{-3mm}
\begin{tabular}{ccccc}
& $\text{NFEs}=15$ & $\text{NFEs}=20$ & $\text{NFEs}=30$ & $\text{NFEs}=50$ \\ 
\rotatebox{90}{\hspace{12mm}{DDIM}}~ &
\includegraphics[width=0.24\textwidth]{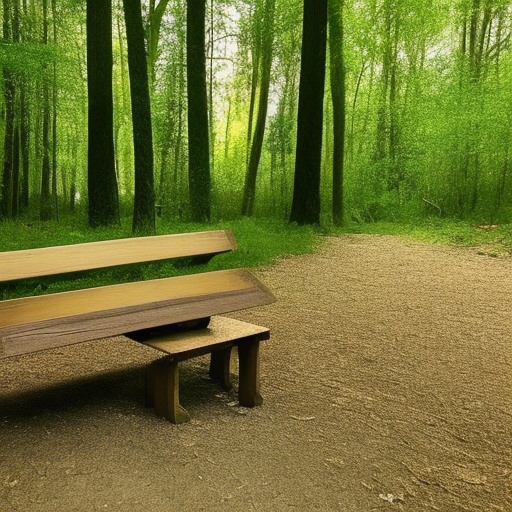} &
\includegraphics[width=0.24\textwidth]{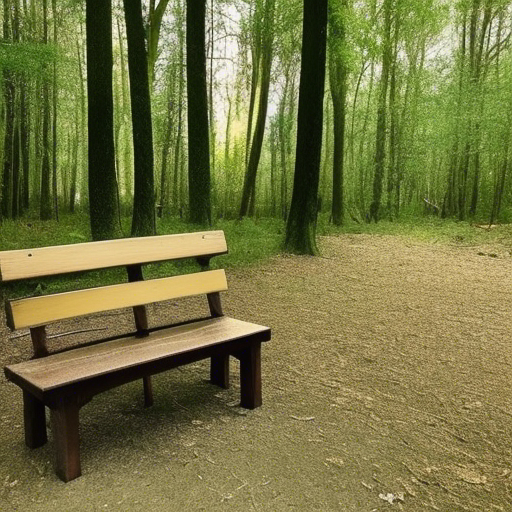} &
\includegraphics[width=0.24\textwidth]{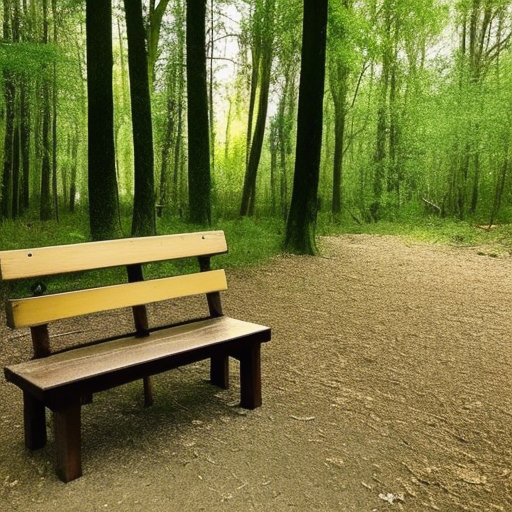} &
\includegraphics[width=0.24\textwidth]{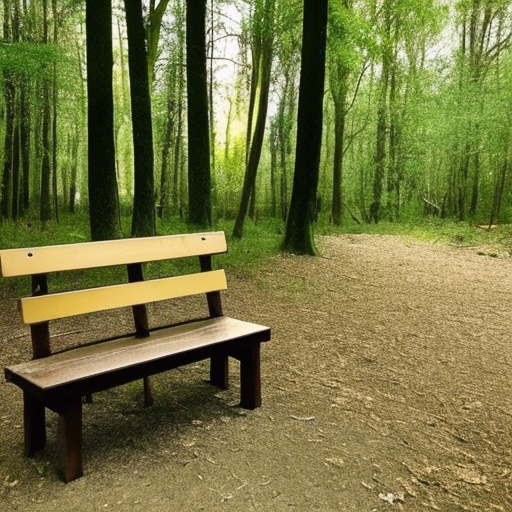} \\
\rotatebox{90}{\hspace{9mm}{RX-DDIM}}~ & 
\includegraphics[width=0.24\textwidth]{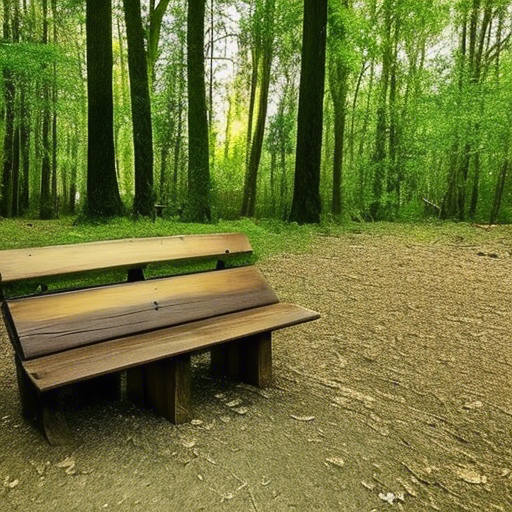} &
\includegraphics[width=0.24\textwidth]{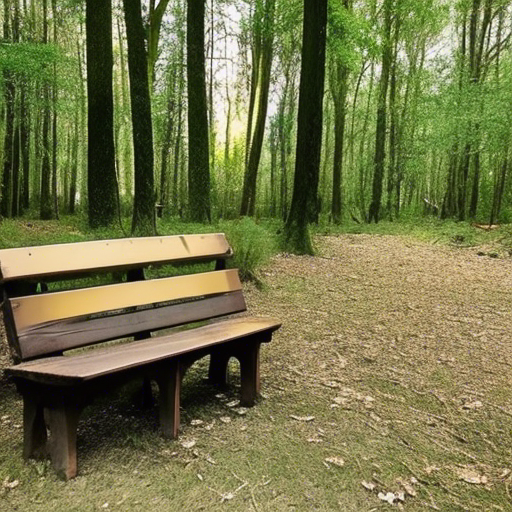} &
\includegraphics[width=0.24\textwidth]{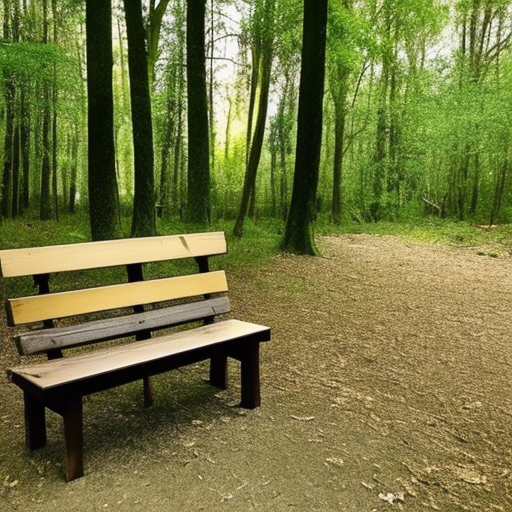} &
\includegraphics[width=0.24\textwidth]{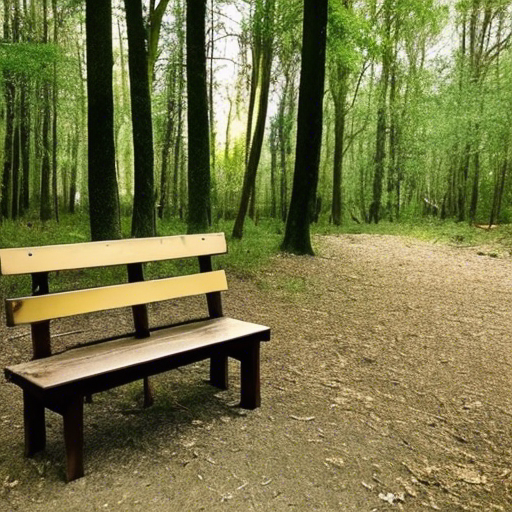} \\
&\multicolumn{4}{c}{\textit{``Small wooden bench sitting in the middle of a forrest.''}}\vspace{1.5mm} \\
\rotatebox{90}{\hspace{12mm}{DDIM}}~ &
\includegraphics[width=0.24\textwidth]{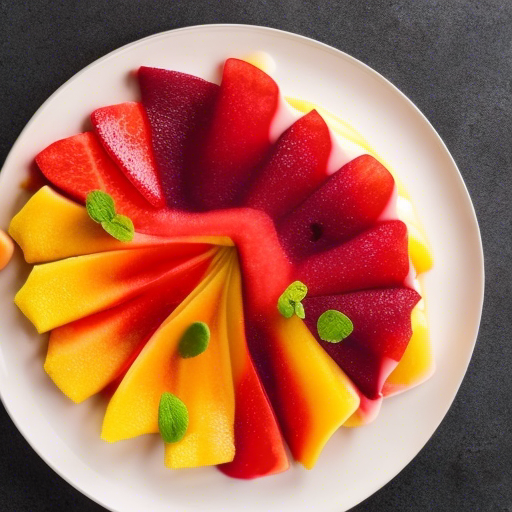} &
\includegraphics[width=0.24\textwidth]{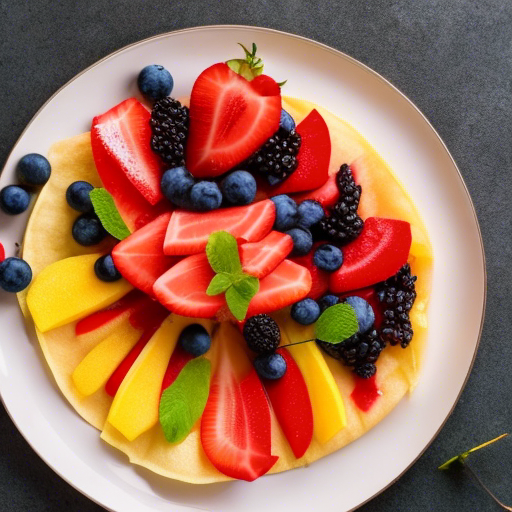} &
\includegraphics[width=0.24\textwidth]{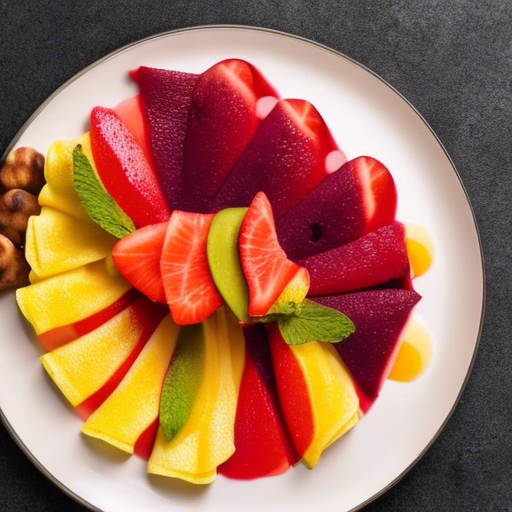} &
\includegraphics[width=0.24\textwidth]{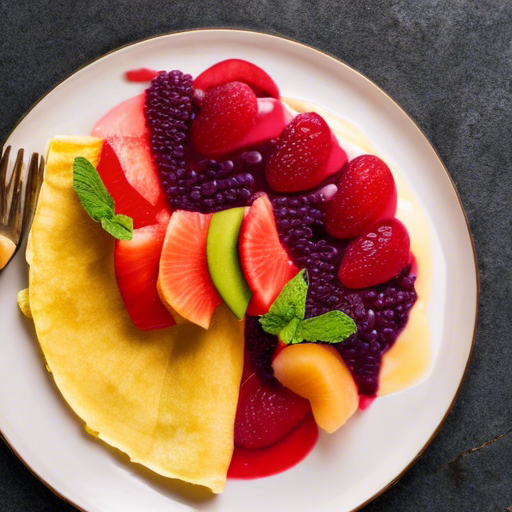} \\
\rotatebox{90}{\hspace{9mm}{RX-DDIM}}~ & 
\includegraphics[width=0.24\textwidth]{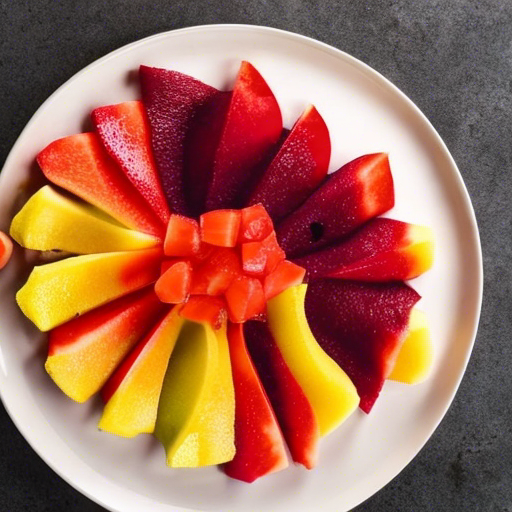} &
\includegraphics[width=0.24\textwidth]{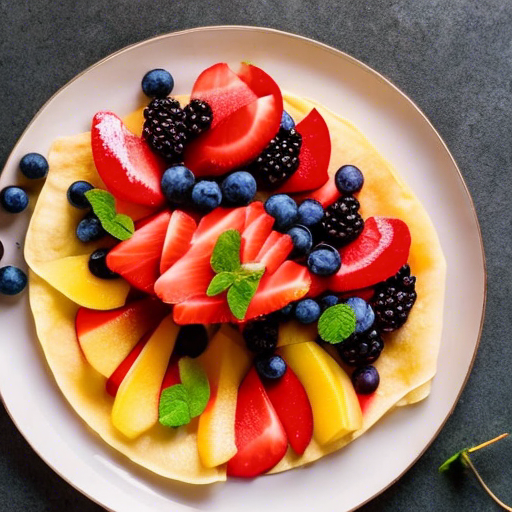} &
\includegraphics[width=0.24\textwidth]{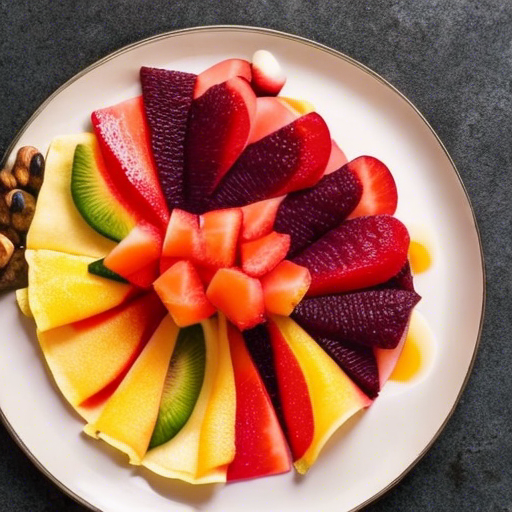} &
\includegraphics[width=0.24\textwidth]{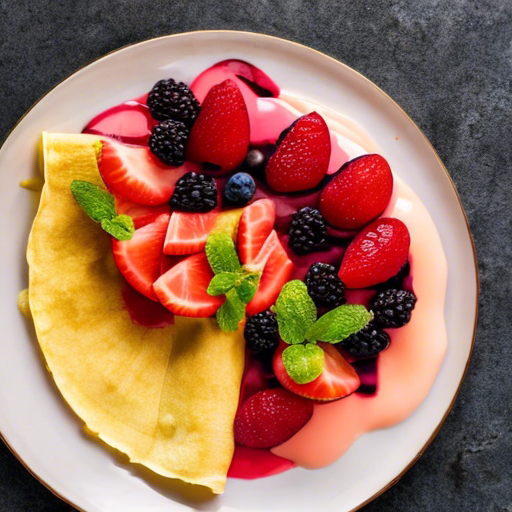} \\
&\multicolumn{4}{c}{\textit{``A plate with  a crepe covered with fresh cut fruit.''}}\vspace{1.5mm} \\
\rotatebox{90}{\hspace{12mm}{DDIM}}~ &
\includegraphics[width=0.24\textwidth]{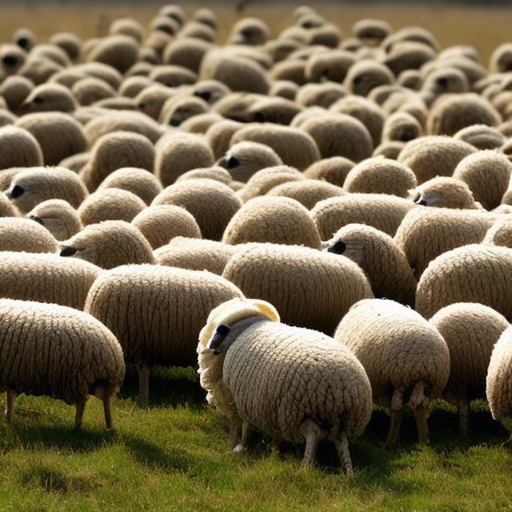} &
\includegraphics[width=0.24\textwidth]{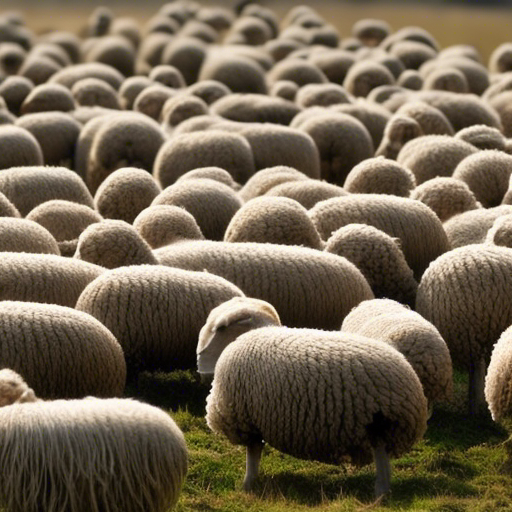} &
\includegraphics[width=0.24\textwidth]{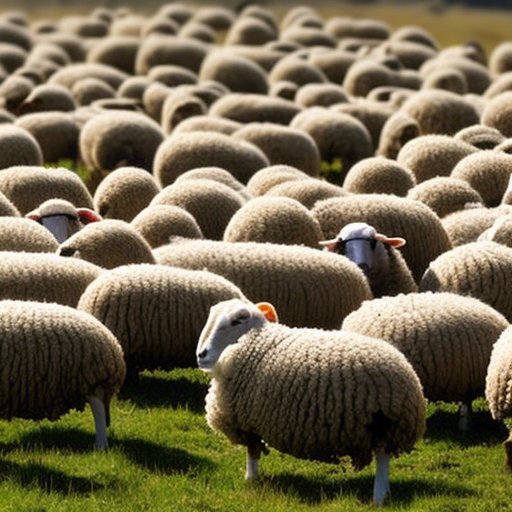} &
\includegraphics[width=0.24\textwidth]{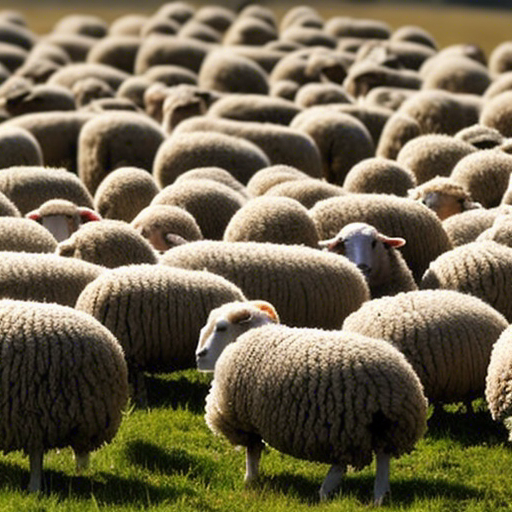} \\
\rotatebox{90}{\hspace{9mm}{RX-DDIM}}~ & 
\includegraphics[width=0.24\textwidth]{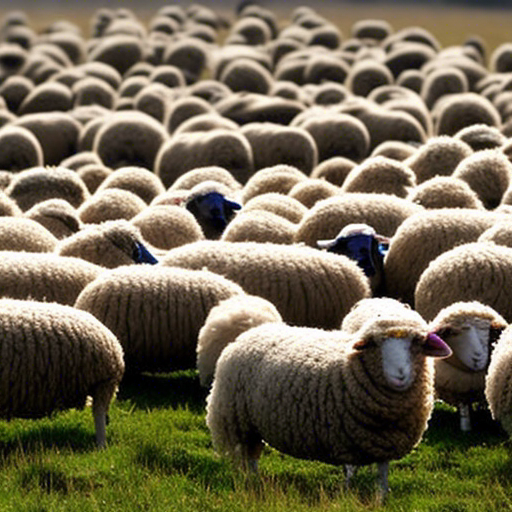} &
\includegraphics[width=0.24\textwidth]{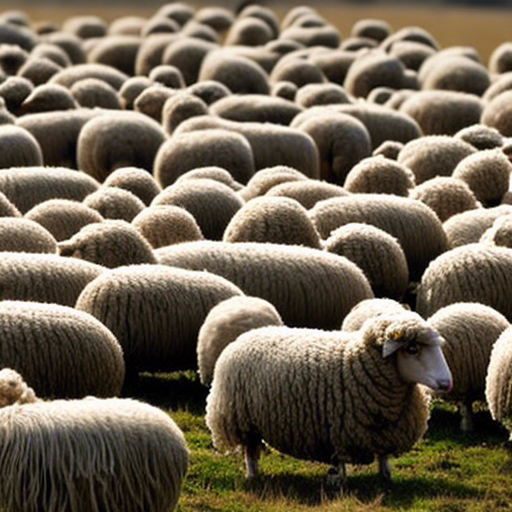} &
\includegraphics[width=0.24\textwidth]{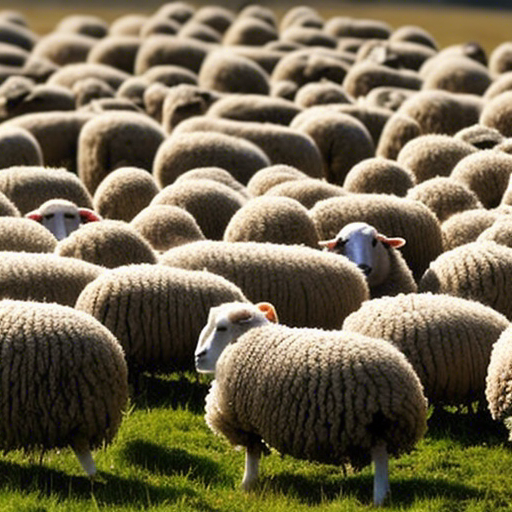} &
\includegraphics[width=0.24\textwidth]{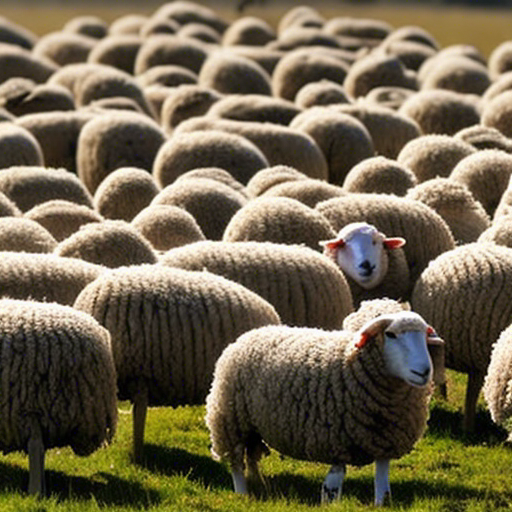} \\
&\multicolumn{4}{c}{\textit{``Very many woolly sheep in a field which is dry.''}}\vspace{1.5mm} 
\end{tabular}
}
\caption{Qualitative results on Stable Diffusion V2 of DDIM and RX-DDIM.}
\label{fig:rx_sd2}
\end{figure}

We provide qualitative results using the EDM backbone in \cref{fig:qual_cifar,fig:qual_ffhq1,fig:qual_ffhq2,fig:qual_afhq1,fig:qual_afhq2,fig:qual_imagenet1,fig:qual_imagenet2} and Stable Diffusion V2 in \cref{fig:rx_sd1,fig:rx_sd2}.
\clearpage


\clearpage

\end{document}